\documentclass{elsarticle}

\usepackage{siunitx}

\usepackage{mycommands}
\usepackage{amsfonts}
\usepackage{lineno}
\usepackage{graphicx}
\usepackage[table]{xcolor}
\usepackage[normalem]{ulem}
\usepackage{subcaption}
\usepackage{booktabs}
\usepackage{tabularx}
\usepackage{hyperref}
\usepackage{xcolor}
\usepackage{chngcntr}
\usepackage{adjustbox}

\journal{Journal of Computers and Chemical Engineering}

\begin{document}

\begin{frontmatter}
\title{Forecasting Industrial Aging Processes with Machine Learning Methods}

%
%

\author[tu]{Mihail Bogojeski}
\author[basf,hmh]{Simeon Sauer}
\author[tu]{Franziska Horn}
\author[tu,kor,mpi]{Klaus-Robert M\"uller\corref{corrauth}}
\ead{klaus-robert.mueller@tu-berlin.de}

\cortext[corrauth]{Corresponding author}
\address[tu]{Machine Learning Group, Technische Universit\"at Berlin, Berlin, 10587, Germany}
\address[basf]{BASF SE, Ludwigshafen, 67056, Germany}
\address[hmh]{Hochschule Mannheim, Mannheim, 68163, Germany}
\address[kor]{Department of Brain and Cognitive Engineering, Korea University, Anam-dong, Seongbuk-gu, Seoul 02841, Korea}
\address[mpi]{Max-Planck-Institut f\"ur Informatik, Saarbr\"ucken, 66123, Germany}



\begin{abstract}
Accurately predicting industrial aging processes makes it possible to schedule maintenance events further in advance, ensuring a cost-efficient and reliable operation of the plant. So far, these degradation processes were usually described by mechanistic or simple empirical prediction models. In this paper, we evaluate a wider range of data-driven models, comparing some traditional stateless models (linear and kernel ridge regression, feed-forward neural networks) to more complex recurrent neural networks (echo state networks and LSTMs). We first examine how much historical data is needed to train each of the models on a synthetic dataset with known dynamics. Next, the models are tested on real-world data from a large scale chemical plant. Our results show that recurrent models produce near perfect predictions when trained on larger datasets, and maintain a good performance even when trained on smaller datasets with domain shifts, while the simpler models only performed comparably on the smaller datasets.
\end{abstract}


\end{frontmatter}


\section{Introduction}\label{sec:intro}
Aging of critical assets is an omnipresent phenomenon in any production environment, causing significant maintenance expenditures or leading to production losses. The understanding and anticipation of the underlying degradation processes is therefore of great importance for a reliable and economic plant operation, both in discrete manufacturing and in the process industry.

With a focus on the chemical industry, notorious aging phenomena include the deactivation of heterogeneous catalysts \cite{forzatti1999catalyst} due to coking \cite{barbier1986deactivation}, sintering \cite{HARRIS1995}, or poisoning \cite{nielsen1995poisoning}; plugging of process equipment, such as heat exchangers or pipes, on process side due to coke layer formation \cite{cai2002coke} or polymerization \cite{wu2018data}; fouling of heat exchangers on water side due to microbial or crystalline deposits \cite{muller2000heat}; erosion of installed equipment, such as injection nozzles or pipes, in fluidized bed reactors \cite{wang1996erosion,werther2000fluidized}; corrosion inside pipes or vessels \cite{Nesic2007}; and more.

Despite the large variety of affected asset types in these examples, and the completely different physical or chemical degradation processes that underlie them, all of these phenomena share some essential characteristics:
\begin{enumerate}
   \item The considered critical asset has one or more key performance indicators (KPIs), which quantify the progress of degradation.\footnote{Such a KPI can either be measured directly or often only indirectly through proxy variables. For example, while catalyst activity is not measured directly in process data, it manifests itself in reduced yield and/or conversion of the process.}
   \item On a time scale much longer than the typical production time scales (i.e., batch time for discontinuous processes; typical time between set point changes for continuous processes), the KPIs drift more or less monotonically to ever higher or lower values, indicating the occurrence of an irreversible degradation phenomenon. (On shorter time scales, the KPIs may exhibit fluctuations that are not driven by the degradation process itself, but rather by varying process conditions or background variables such as, e.g., the ambient temperature.)
   \item The KPIs return towards their baseline after maintenance events, such as cleaning of a fouled heat exchanger, replacement of an inactive catalyst, etc.
   \item The degradation is no `bolt from the blue' -- such as, e.g., the bursting of a flawed pipe --, but is rather driven by creeping, inevitable wear and tear of process equipment.
\end{enumerate}
Any aging phenomenon with these general properties is addressed by the present work.

Property (4) suggests that the evolution of a degradation KPI is to a large extent determined by the process conditions, and not by uncontrolled, external factors. This sets the central goal of the present work: To forecast the evolution of the degradation KPI  over a certain time horizon, given the planned process conditions in this time frame. If instead external factors such as humidity, ambient temperature, human interventions, etc.\ are the dominating driving forces of an aging process, the presented approach is not applicable.

For virtually any important aging phenomenon in chemical engineering, the respective scientific community has developed a detailed understanding of their microscopic and macroscopic driving forces. This understanding has commonly been condensed into sophisticated  mathematical models. 
Examples of such mechanistic degradation models deal with coking of steamcracker furnaces \cite{GAO2009501,DeSchepper2010,BERRENI20112876}, sintering \cite{ruckenstein1973growth, li2017modeling} or coking \cite{froment2001modeling} of heterogeneous catalysts, or crystallization fouling of heat exchangers \cite{brahim2003numerical}.

While these models give valuable insights into the dynamics of experimentally non-accessible quantities, and can help to verify or falsify hypotheses about the degradation mechanism \textit{in general}, they are usually not (or only with significant modeling effort) transferable to the \textit{specific} environment in a real-world apparatus: Broadly speaking, they often describe `clean' observations of the degradation process in a lab environment, and do not reflect the `dirty' reality in production, where additional effects come into play that are hard or impossible to model mechanistically. To mention only one example, sintering dynamics of supported metal catalysts are hard to model quantitatively even in the `clean' system of Wulff-shaped particles on a flat surface \cite{li2017modeling} -- while in real heterogeneous catalysts, surface morphology and particle shape may deviate strongly from this assumption.
Another disadvantage of mechanistic models is that their numerical solution can be computationally expensive or even intractable.

While the latter issue of computational complexity can be mitigated by surrogate modeling methods \cite{wang2018constrained,kim2020surrogate}, the former issue of real-world complexity is the main reason why mechanistic models of degradation dynamics are rarely used in a production environment.

This weakness is addressed by hybrid process models (also referred to as gray-box models), which combine mechanistic with data driven modeling approaches to bridge the gap between idealized mechanistic models and the real world \cite{von2014hybrid,willis2017simultaneous,asprion2019gray,zendehboudi2018applications,glassey2018hybrid}. However, to the best of our knowledge, these models have not yet been used to forecast industrial aging processes in chemical plants.

When dealing with the complexity of modeling real-world aging processes, statistical approaches have proven successful in a variety of applications. For example, data-driven methods for fault detection of chemical plants \cite{russell2012data}, such as multivariate anomaly detection with fisher discriminant \cite{chiang2000fault} or principal component analysis \cite{Kresta1991pca}, are routinely applied nowadays to monitor process equipment. However, we emphasize that most of these applications focus on the \textit{detection} or \textit{monitoring} of degradation, not on the \textit{prediction} of its progression.

Publications of data-driven models that \textit{predict} degradation dynamics predominantly address specific aging phenomena, such as batch-to-batch fouling of heat exchangers as a function of the polymer type produced in the respective batch \cite{wu2018data}, or fouling of the crude preheat train in petroleum refineries \cite{radhakrishnan2007heat,aminian2008evaluation}.
To the best of our knowledge, all published work in this field is either based on classical statistical regression methods, such as ordinary or partial least squares \cite{wu2018data}, or on small-scale machine learning (ML) methods, such as small feed-forward neural networks (FFNN) trained with limited datasets \cite{radhakrishnan2007heat,aminian2008evaluation}.
So far, advanced ML algorithms, such as recurrent neural networks (RNN), trained with years or even decades of historical plant data, have not been studied in depth in the context of predicting degradation of chemical process equipment \cite{LEE2018111}. It is the aim of this work to investigate the prospects of advanced ML methods for this problem, compare them to classical regression methods and understand potential limitations.

The rest of this paper is structured as follows: First, we formalize the general IAP problem setting (Section~\ref{sec:problem_setting}). Then we describe our two datasets (Section~\ref{sec:data}), as well as quickly introduce the five ML models that we evaluated for this task (Section~\ref{sec:ml_methods}). Finally, we present the prediction results of the different models on both datasets (Section~\ref{sec:results}) and conclude the paper with a discussion (Section~\ref{sec:disc}).

\section{Problem Definition}\label{sec:problem_setting}

\begin{figure}[htb!]
  \centering
    \includegraphics[width=\textwidth]{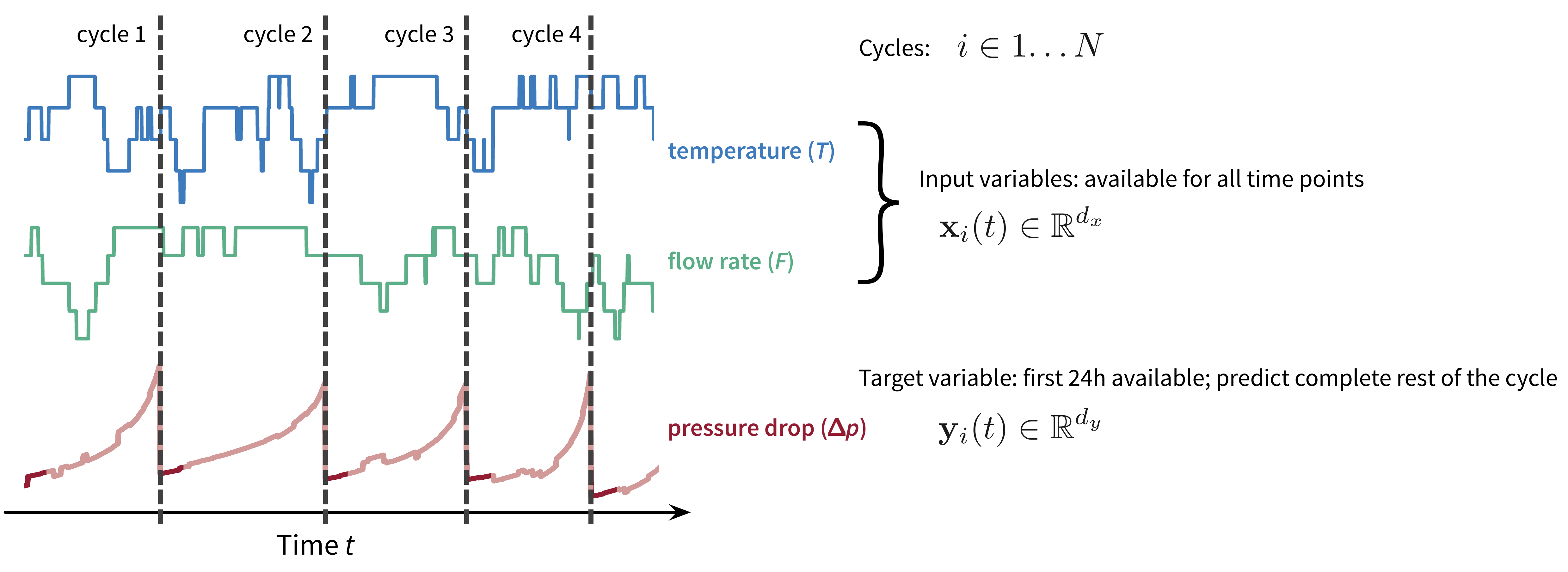}
    \caption{Illustration of the industrial aging process (IAP) forecasting problem. The degradation KPI (e.g.\ pressure drop $\Delta p$ in a fixed bed) increases over time (e.g.\ due to coking), influenced by the (manually controlled) process conditions (e.g.\ reaction temperature $T$ and flow rate $F$). The KPI recovers after a maintenance event, which segments the time axis into different degradation cycles. The IAP forecasting task is to predict the evolution of the KPI, i.e., the target (dependent) variable $\mathbf{y}_i(t)$, in the current cycle $i$, given the upcoming process conditions, i.e., the input (independent) variables $\mathbf{x}_i(t)$.}
    \label{fig:IAP_problem}
\end{figure}

The general industrial aging process (IAP) forecasting problem is illustrated in Fig.~\ref{fig:IAP_problem}: The aim is to model the evolution of one or several degradation KPIs $\y(t)$  as a function of the planned process conditions $\x(t)$ over the time frame of an entire degradation cycle, i.e., until the next required maintenance event. Formally, the problem statement reads:

\paragraph*{Given} A time window $t \in [0, T_i]$ between two maintenance events, referred to as the $i$-th degradation cycle, and the planned process conditions $\x_i(t) \in \R^{d_x}$ within this time window.

\paragraph*{Determine} An estimate $\hat{\y}_i(t) \in \R^{d_y}$ of the evolution of one or several degradation KPIs $\y_i(t)$ in the $i$-th cycle.

\paragraph*{Objective} Minimal discrepancy between the estimated and true KPI values, as measured by mean squared error (MSE), cf.\ Eq.~\eqref{eq:MSE}.

Some peculiarities inherent to IAPs make this problem more challenging than it first appears.
First, degradation phenomena can exhibit pronounced memory effects, which means that a certain input pattern $\x(t)$ may affect the output $\y(t')$ only at much later times $t' > t$. In addition, these memory effects can also occur across multiple time scales, which makes these processes notoriously hard to model.
As an example, consider a heat exchanger suffering from coking of the inner tube walls. The observed heat transfer coefficient serves as KPI $\y_i(t)$, and process conditions $\x_i(t)$ comprise mass flow, chemical composition and temperature of the processed fluid. The time horizon is one cycle between two cleaning procedures (e.g.\ burn-off). If at an early time $t_1$ in the cycle an unfavorable combination of low mass flow, high content of coke precursors, and high temperature occurs, first coke patches can form at the wall, which are not yet big enough to impact heat transfer significantly. However, they serve as a nuclei for further coke formation later in the cycle, so that $\y_i(t)$ drops faster at $t>t_1$ compared to a cycle where the process conditions were not unfavorable around $t_1$, but with very similar process conditions throughout the rest of the cycle.

An additional complication arises from the fact that in real application cases, the distinction between degradation KPI $\y$, process conditions $\x$, and uncontrolled influencing factors is not always clear-cut. Consider, for example, the case of a heterogeneous catalyst subject to deactivation, where the loss of catalytic activity leads to a decreased conversion rate. In this case, the conversion rate could serve as a target degradation KPI $\y$, while process conditions, such as the temperature, which are manually controlled by the plant operators, would be considered input variables $\x$ for the model. However, the plant operators might try to keep the conversion rate at a certain set point, which can be achieved by raising the temperature to counteract the effects of the catalyst degradation. This introduces a feedback loop between the conversion rate and the temperature, which means the temperature can not be considered an independent variable anymore, as its actual value (partially) depends on the target. Therefore, care has to be taken, since including such a dependent variable as an input $\x$ in a model could lead one to report overly optimistic prediction errors that would not hold up when the model is later used in reality.

\section{Datasets}\label{sec:data}
To gain insights into and evaluate different ML models for the IAP forecasting problem, we consider two datasets: one synthetic, which we generated ourselves using a mechanistic model, and one containing real-world data from a large plant at BASF. Both datasets are described in more detail below.

The reason for working with synthetic data is that this allows us control two important aspects of the problem: data \textit{quantity} and data \textit{quality}. Data quantity is measured, e.g., by the number of catalyst lifecycles in the dataset, which can be chosen as large as we want for synthetic data, to test even the most data-hungry ML methods. Data quality refers to the level of noise in the dataset, or, in other words, the degree to which the degradation KPI $\y(t)$ is uniquely determined by the provided process conditions $\x(t)$ in the dataset. In a synthetic dataset based on a deterministic degradation model, we \textit{know} that there is a functional mapping between $\x$ and $\y$, i.e., there exists no fundamental reason that could prevent a ML model from learning this relation with vanishing prediction errors. In contrast, with real data, a bad prediction error can either be a problem of the method, and/or of the dataset, which might not contain sufficient information on the input side $\x$ to accurately predict the output quantity $\y$.

Please note that the synthetic dataset refers to a different chemical process than the real-world dataset. As a consequence, models trained on the synthetic data have to be re-trained when applying them to the real-world dataset.

\subsection{Synthetic dataset}\label{sec:SyntheticData}
For the synthetic dataset, we modeled the wide-spread phenomenon of slow, but steady loss of catalytic activity in a continuously operated fixed-bed reactor. Ultimately, the catalyst deactivation leads to unacceptable conversion or selectivity rates in the process, necessitating a catalyst regeneration or replacement, which marks the end of one cycle.

The chemical process in the reactor under consideration is the gas-phase oxidation of an olefine. To generate the time series for all variables, we used a mechanistic process model with the following ingredients (further details can be found in Section~\ref{ssub:appdx_synth_model} in the supplement):
\begin{itemize}
\item Mass balance equations for all five relevant chemical species (olefinic reactant, oxygen, oxidized product, CO$_2$, water) in the reactor, which is, for simplicity, modeled as an isothermal plug flow reactor, assuming ideal gas law. The reaction network consists of the main reaction (olefine + O$_2$ $\rightarrow$ product) and one side reaction (combustion of olefine to CO$_2$).
\item A highly non-linear deactivation law for the catalyst activity, which depends on reaction temperature, flow rate, and inflowing oxygen, as well as the activity itself.
\item Kinetic laws for the reaction rates.
\item A stochastic process determining the process conditions (temperature, flow rate, etc.).
\end{itemize}
Based on the current process conditions and hidden states of the system, we used the mechanistic model to generate a multivariate time series $[\x(t), \y(t)]$ comprising 50 years of historical data, separated into 2153 degradation cycles with a total of 435917 time points. The final dataset consists of six input features $\x(t)$ at each time point $t$, namely the five process parameters (mass flow rate, reactor pressure, temperature, and mass fractions of the two reactants olefine and O$_2$) and the time since the last maintenance event, while the two degradation KPIs (conversion and selectivity) constitute the target variables $\y(t)$ (Table~\ref{tab:syndat_vars}).
\begin{table}[htbp!]
  \centering
    \caption{Variables in the synthetic dataset used for machine learning. The variable type indicates if a quantity is an input (x) or output (y) in the IAP forecasting problem.}
    \label{tab:syndat_vars}
    \begin{tabular}{cclc}
      Variable Name & Unit & Description & Type \\
      \hline
          $p$ & mbar & reactor pressure set point& x \\
          $T$ & ${}^\circ$C & reaction temperature &  x \\
          $F$ & kg/h & total mass flow into reactor &x  \\
          $\mu_\textrm{Olef.}^{(in)}$ & \% & mass fraction of olefine at inlet& x \\
          $\mu_{\textrm{O}_2}^{(in)}$ & \% & mass fraction of oxygen at inlet& x \\
          $t$ & h & time-on-stream, i.e., hours since last regeneration & x \\
          $C$ & \% & Conversion & y \\
          $S$ & \% & Selectivity & y \\
    \end{tabular}
\end{table}

\begin{figure}[htb!]
  \centering
    \includegraphics[width=\textwidth]{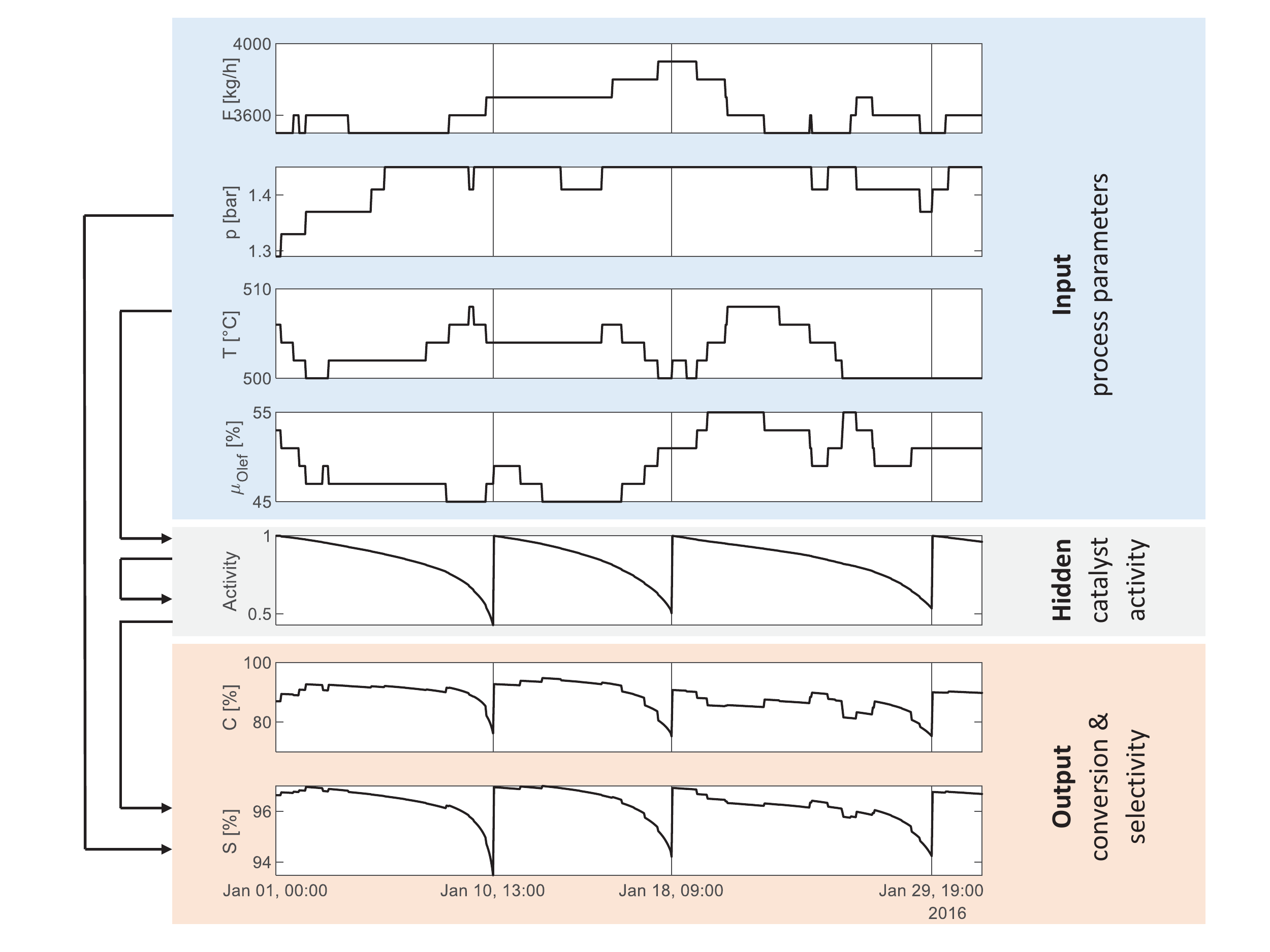}
    \caption{One month of the synthetic dataset, showing loss of catalytic activity in a fixed-bed reactor. At each time point $t$, the vector of process conditions $\x(t)$ includes the reactor temperature $T$, mass flow rate $F$, reactor pressure $p$, and mass fractions $\mu_z$ of the reactants at the reactor inlet. Degradation KPIs $\y(t)$ are conversion and selectivity of the process.}
    \label{fig:desc_data_syn}
\end{figure}

To give an impression of the simulated time series, one month of data is shown in Fig.~\ref{fig:desc_data_syn}. The duration of deactivation cycles is around 8-10 days. The catalyst activity $A(t)$ is a hidden state and therefore not part of the dataset, but is only shown to illustrate the dynamics of the problem: System output $\y(t)$ (selectivity and conversion) is not only affected by the current process parameters $\x(t)$, but also the current catalyst activity $A(t)$, which is non-linearly decreasing over each cycle.

\subsection{Real-world dataset}
The second dataset contains process data for the production of an organic substance in a continuous world-scale production plant at BASF. The process is a gas phase oxidation in a multi-tubular fixed-bed reactor.

The catalyst particles in the reactor suffer from coking, i.e., surface deposition of elementary carbon in the form of graphite. This leads to reduced catalytic activity and increased fluid resistance. The latter is the more severe consequence and leads to an increasing pressure drop over the reactor, as measured by the difference $\Delta p$ of gas pressure before and after the reactor. The physical reason for the increasing pressure drop is the reduction of void fraction in the catalyst bed due to the build-up of coke residues, as described by Ergun's equation \cite{Ergun1952} and its extensions \cite{Nemec2005}.

\begin{figure}[htb!]
  \centering
    \includegraphics[width=\textwidth]{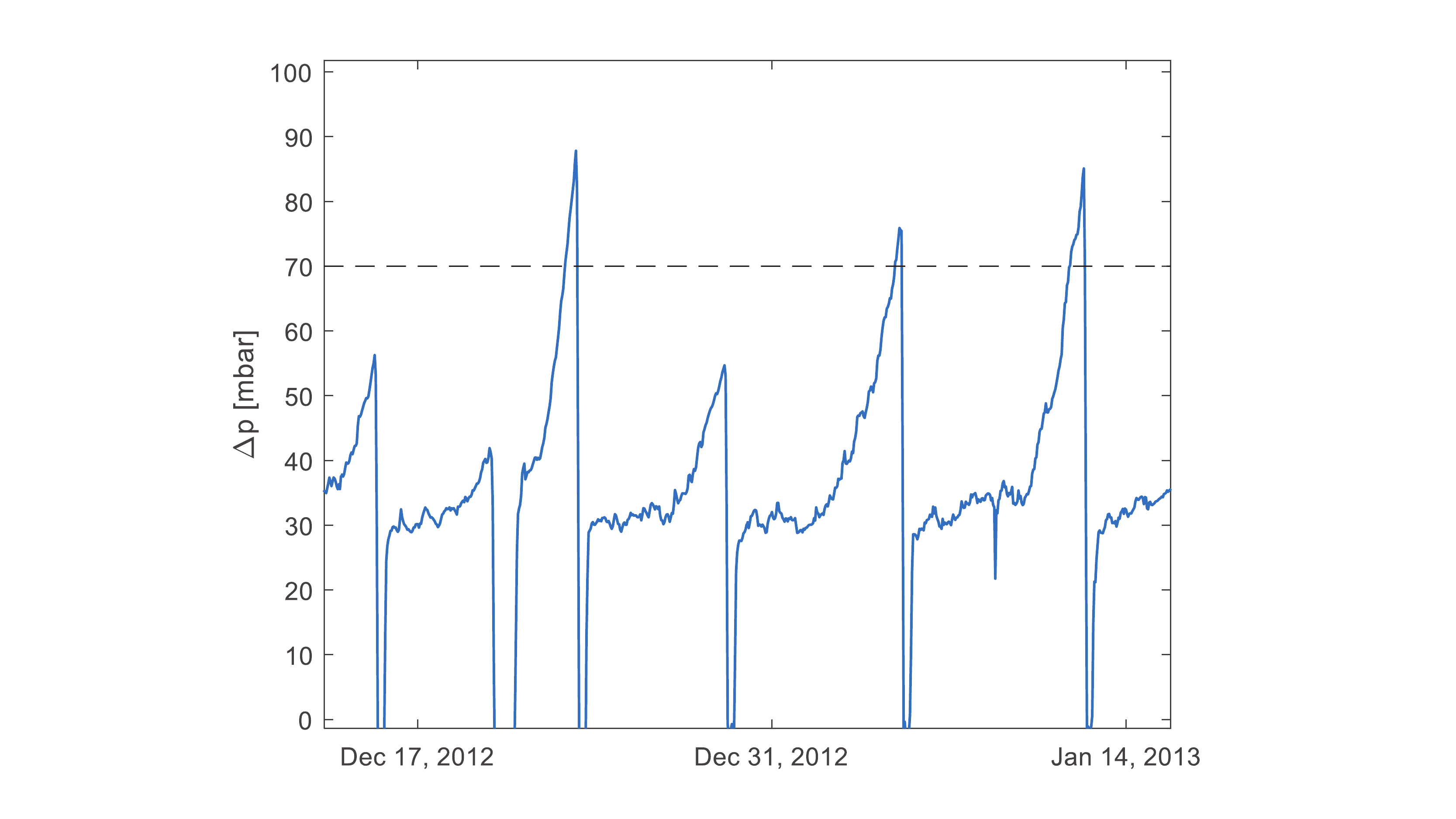}
    \caption{One month of historic data of the real-world dataset, showing the pressure loss $\Delta p$ over the reactor, which is the degradation KPI $\y(t)$ in this IAP forecasting problem. When $\Delta p$ reaches a value in the order of the EOR threshold of 70 mbar, the cokes deposit is burned off, which marks the end of a cycle.}
    \label{fig:desc_data_real}
\end{figure}

When $\Delta p$ exceeds a pre-defined threshold, the so-called end-of-run (EOR) criterion is reached. Then, the coke layer is burned off in a dedicated regeneration procedure, by inserting air and additional nitrogen into the reactor at elevated temperatures for a variable number of hours. Operational reasons can lead to a delayed burn-off with $\Delta p$ exceeding the EOR threshold, or, vice versa, a premature burn-off when $\Delta p$ has not yet reached the EOR threshold. Some exemplary cycles for $\Delta p$ are shown in Fig.~\ref{fig:desc_data_real}.

Since coke is not removed perfectly by this burn-off procedure, coke residues accumulate from regeneration to regeneration, making the pressure drop issue ever more severe. Therefore, the entire catalyst bed must be replaced every 6-24 months.

Suspected influencing factors for the coking rate are:
\begin{enumerate}
\item mass flow $F$ through the reactor (``feed load'')
\item ratio of organic reactant to oxygen in the feed
\item intensity of previous regeneration procedures
\item length of the previous degradation cycle

\end{enumerate}

The dataset includes seven years of process data from the four most relevant sensors, extracted from the plant information management system (PIMS) of the plant. Given the time scale of 4 to 7 days between two burn-off procedures, this corresponds to 375 degradation cycles belonging to three different catalyst batches. The sampling rate is 1/hour for all variables with a linear interpolation to that time grid. After removing some outlier cycles (shorter than 50 hours), the final size of the dataset is 327 cycles for a total of 36058 time points.

The task is to predict, at an intermediate moment $t_k$ during a degradation cycle, the coking-induced pressure drop $\Delta p$ over the entire remaining duration of the cycle. Of particular interest is a prediction of the time point $t_\textrm{EOR}$ at which the EOR threshold $\Delta p^\textrm{max}=70\,\text{mbar}$ is reached.
The input variables $\x(t)$ of the model consist of the process parameters, as well as several engineered features (Table~\ref{tab:realdat_feat}), either derived from the process parameters themselves or from the degradation KPI $\Delta p$ in the previous cycles, yielding a total of 10 input features.\footnote{The latter features are in particular relevant as they may encode information about the long-term effects in the system, such as coke residues accumulating on the time scale of months and years. Forecasting these long-term effects is not the subject of the IAP forecasting problem for this dataset; we rather focus on forecasting the currently running cycle. Therefore, by the time the forecast is generated, the previous cycle has already been observed and may be used for feature engineering.}
\begin{table}[htbp!]
  \centering
  \caption{Variables in the real-world dataset used for machine learning. The variable type indicates if a quantity is an input (x) or output (y) in the IAP forecasting problem. The first four input variables are the process parameters (measured with sensors in the plant), all others are additional engineered features. The total inflowing reactant F\_R is a mixture of fresh reactant from a tank reservoir (F\_FRESH), and recycled reactant from downstream separation units. The variable F\_FRESH is not direclty used as an input feature, but is necessary for the definition of F\_FRESH / F\_R.}
  \label{tab:realdat_feat}
  \begin{adjustbox}{width=\columnwidth, center}
    \begin{tabular}{cclc}
      Variable Name & Unit & Description & Type \\
      \hline
          T & ${}^\circ$C & reaction temperature &  x \\
          F\_R & kg/h & inflow of organic reactant into reactor &x  \\
          F\_AIR & kg/h & mass inflow air into reactor & x \\
          F\_FRESH & kg/h & fresh organic reactant in inflow & x \\
          cycle\_no & - & counter to index different cycles & x \\
          t\_react & h & duration of current cycle & x\\
          last\_PD & mbar & pressure difference at the end of the previous cycle & x\\
          F\_R/(F\_R+F\_AIR) & \% & mass fraction of reactant in inflow & x\\
          F\_FRESH / F\_R & \% & fraction of fresh organic reactant in inflow & x\\
          F\_CUM\_CYCLE & kg & F\_R cumulated over running cycle & x \\
          F\_CUM\_CAT & kg & F\_R cumulated over catalyst lifetime & x \\
          PD & mbar & pressure difference $\Delta p$ over reactor & y \\
    \end{tabular}
  \end{adjustbox}
\end{table}

\section{Machine Learning Methods}\label{sec:ml_methods}
We now frame the IAP forecasting problem described in Section~\ref{sec:problem_setting} as a machine learning problem, by defining a concrete function $f$ that returns $\hat{\y}_i(t)$, an estimate of the KPIs at a time point $t$ in the $i$-th degradation cycle, based on the process conditions $\x_i$ at this time point as well as possibly up to $k$ hours before $t$:
\begin{linenomath*}\begin{align}\label{eq:problem_setting}
  \hat{\y}_i(t) = f\left(\x_i(t) \left[, \x_i(t-1), \dots, \x_i(t-k)\right]\right)\quad \forall t \in [0, \dots, T_i].
\end{align}\end{linenomath*}
The task is to predict $\y_i(t)$ for the complete cycle (i.e., up to $T_i$), typically starting from about 24 hours after the last maintenance event that concluded the previous cycle.\footnote{In Eq.~\eqref{eq:problem_setting}, the prediction function $f$ is defined as a function of the current and past input variables $\x_i$. Since usually the values of the degradation KPIs $\y_i$ are known for at least the first 24 hours of each cycle, in principle the set of input variables of $f$ could be extended to also include $\y_i(t')$ for $t' < t$. However, while this might improve the predictions at the beginning of the cycle, since our aim is to predict the complete cycle starting after the first 24 hours, for the predictions for most time points, not the real values $\y_i(t')$ could be used as an input, but instead their predicted values $\hat{\y}_i(t')$ would have to be used. Since these predicted values typically contain at least a small error, the forecast for time points further in the future would be based on noisier and noisier input data, as the prediction errors in the input variables $\hat{\y}_i(t')$ would quickly accumulate. Therefore, the only explicit inputs to the model are the predefined process conditions $\x_i$.}
For simplicity, in many cases we only write $\x$ and $\y$, omitting the reference to the current cycle $i$ and time points $t$ in questions, while $\x$ might include the process conditions for multiple time points from a fixed time window in the past (i.e.\ up to $t-k$).

\begin{figure}[htb!]
  \centering
  \begin{subfigure}[b]{0.39\textwidth}
    \caption{}
    \vspace{-5pt}
    \includegraphics[width=\textwidth]{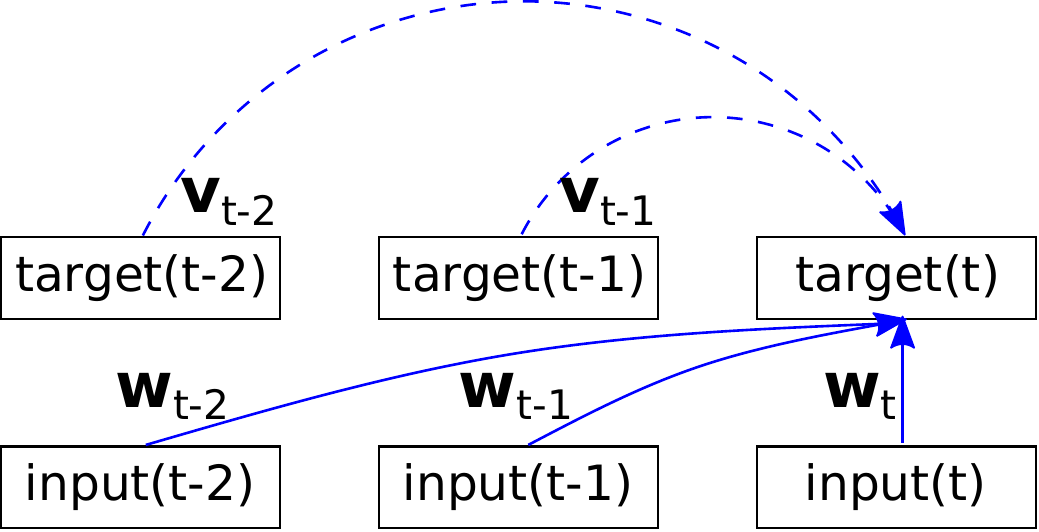}
  \end{subfigure}
  \hspace{10pt}
  \begin{subfigure}[b]{0.45\textwidth}
    \caption{}
    \vspace{10pt}
    \includegraphics[width=\textwidth]{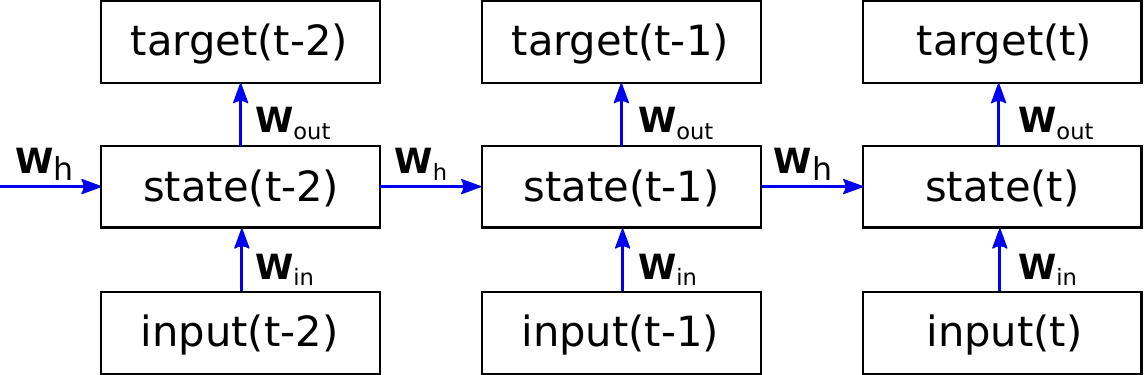}
  \end{subfigure}
  \caption{A comparison of stateless and stateful models for time series forecasting. \textbf{(a)} shows a
  stateless model, which bases the predictions on the information contained in a fixed time window in the
  past, while \textbf{(b)} displays a stateful model, where information about the past is maintained and
  propagated using a hidden state.}
  \label{fig:stateful_v_stateless}
\end{figure}

In this paper we examine five different machine learning models, which can be divided into two main subgroups: stateless and stateful models (Fig.~\ref{fig:stateful_v_stateless}).

Stateless models directly predict the target given the inputs from a fixed time window in the past, independent of the predictions for previous time points. The stateless models used in this paper are linear ridge regression (LRR), kernel ridge regression (KRR), and feed-forward neural networks (FFNN), i.e., one linear and two non-linear prediction models. Stateful models, on the other hand, maintain an internal hidden state of the system that encodes information about the past and which is utilized in addition to the current process conditions when making a prediction. The stateful models we explore in this paper are two recurrent neural network (RNN) models~\cite{mandic2001recurrent}: echo state networks (ESN) and long short term memory (LSTM) networks.
The five ML models are briefly introduced in the following subsections, while a more detailed description can be found in Section~\ref{ssec:mldetail} in the supplement.

As the models utilize the inputs differently when making their predictions, the kind of machine learning method that is chosen for the forecasting task determines the exact form of the function $f$ from Eq.~\eqref{eq:problem_setting}. Yet, while the functional form of $f$ is predetermined, its exact parameters need to be adapted to fit the dataset at hand in order to yield accurate predictions. For this, the available data is first split into training and test sets, where each of the two sets contains the entire multivariate time series from several mutually exclusive degradation cycles from the original dataset, i.e., multiple input-output pairs $\left\{\x_i(t), \y_i(t)\right\}_{t \in [0, T_i]}$ consisting of the planned conditions $\x$ and degradation KPIs $\y$ of the given process. Then, using the data in the training set, the ML algorithm learns the optimal parameters of $f$ by minimizing the expected error between the predicted KPIs $\hat{\y}_i(t)$ and the true KPIs $\y_i(t)$~\cite{vapnik1995nature, muller2001introduction, hansen2013assessment}. After the ML model has been trained, i.e., when $f$ predicts $\y_i(t)$ as accurately as possible on the training set, the model is evaluated on the test set, i.e., previously unseen data, to give an indication of its performance when later used in reality. If the performance on the training set is much better than on the test set, the model does not generalize well to new data and is said to have ``overfit'' on the training data.

In addition to the regular parameters of $f$, many ML models also require setting some hyperparameters, that, for example, determine the degree of regularization (i.e., how much influence possible outliers in the training set can have on the model parameters).
The values for these hyperparameters are determined by training models with different settings on a subset of the training data and evaluating them on the remaining data from the training set (also called the validation set) and then choosing those hyperparameters with the best performance on this validation split. The final model with the selected hyperparameters is then trained on the complete training set and evaluated on the test set as described above. Specific details about the hyperparameters of the different models and their chosen values for each dataset can be found in Section~\ref{ssec:hyperparams} in the supplement.

\subsection{Linear ridge regression (LRR)}

LRR is based on the standard linear regression model, but with an added regularization term that prevents the weights from taking on extreme values due to outliers in the training set. The target variables $\y$ are predicted as a linear combination of the input variables $\x$, i.e.,
\begin{linenomath*}\begin{align*}
\hat{\y} = \vW\x,
\end{align*}\end{linenomath*}
where $\vW \in \mathbb{R}^{d_y \times d_x}$ is a weight matrix, i.e., the model parameters of $f$ that are learned from the training data.
The simple model architecture, globally optimal solution, and regularization of LRR all contribute to reducing overfitting of the model. Additionally, training and evaluating the model is computationally cheap, making it a viable model for large amounts of data as well.

The regularization parameter $\lambda$ is the only hyperparameter of LRR that needs optimization, which we select by searching for the best performing hyperparameter on a grid from 0.001 to 10 using a 10-fold cross-validation~\cite{stone1974cross}.

For the synthetic dataset, the input vector at time point $t$ consists of the process
parameters for the past 24 hours, giving the model a time window into the past, i.e. $\x_{24h}(t) = [\x(t);\x(t-1);\dots;\x(t-24)]$. For the real-world dataset, the input at time point $t$ only consists of the process conditions at that time point $\x(t)$, as a larger time window would reduce the number of data points in the training set (if $k$ hours from the past are taken, the inputs for each cycle have to start $k$ hours later, leading to the loss of $k$ samples per cycle) while increasing the number of input features, therefore making overfitting more likely.

\subsection{Kernel ridge regression (KRR)}

KRR is a non-linear regression model that can be derived from LRR using the so called
`kernel trick' \cite{scholkopf1998nonlinear, muller2001introduction, vapnik1995nature, scholkopf2002learning, muller1997predicting}. Instead of using the regular input features $\x$, the features are
mapped to a high (and possibly infinite) dimensional space using a feature map $\phi$, corresponding to some kernel function $k$ such that
$\phi(\x)^T\phi(\x') = k(\x, \x')$.
By computing the non-linear similarity $k$ between a new data point $\x$ and the training examples $\x_j$ for $j = 1, \dots, N$, the targets $\y$ can be predicted as
\begin{linenomath*}\begin{align*}
  \hat{\y} = \sum_{j=1}^N \alpha_{j}k(\x, \x_{j}),
\end{align*}\end{linenomath*}
where $\alpha_j$ are the learned model parameters.\\
The non-linear KRR model can adapt to more complex data compared to LRR while the globally optimal solution can still be obtained analytically, which has made KRR one of the most commonly used non-linear regression algorithms in the field of machine learning~\cite{rasmussen2006regression, smola2003bayesian}. However, the performance of the model is also more sensitive to the choice of hyperparameters, so a careful selection
and optimization of the hyperparameters is necessary. Additionally, the fact that solving the optimization problem scales cubically in runtime and quadratically in memory with the number of training examples $N$ makes it difficult to apply KRR to problems with large training sets.
The scaling issue is exactly one of the challenges we encountered when applying KRR to the synthetic dataset. The linear algebra library we used only supported operations with matrices of a maximum size of 50 000, so when applying KRR to the synthetic dataset we only used 10 percent of the training set, consisting of around 40 000 points.
The KRR model has two hyperparameters that need to be optimized, the regularization parameter $\lambda$ and the kernel width $\sigma$. The optimal hyperparameters were selected with a 10-fold cross-validation from an exponential grid, with values ranging from $10^{-16}$ to 1 for $\lambda$ and from 0.1 to 1000 for $\sigma$.

Analogously to LRR, the KRR model for the synthetic dataset was trained using inputs from a 24-hour time window into the past, while the model for the real-world dataset only used inputs at the current time point.

\subsection{Feed-forward neural networks (FFNN)}

FFNNs were the first and most straightforward neural network architecture to be conceived, yet, due to their flexibility, they are still
successfully applied to many different types of machine learning problems~\cite{bishop1995neural, goodfellow2016deep, schmidhuber2015deep}.
Analogously to LRR, FFNNs learn a direct mapping $f$ between some input features
$\x$ and some output values $\y$. However, unlike a linear model, FFNNs can approximate also highly non-linear
dependencies between inputs and outputs. This is achieved by transforming the input using a succession of
``layers'', where each layer is usually composed of a linear transformation with a weight matrix $\vW$ followed by a non-linear operation $\sigma$:
\begin{linenomath*}\begin{align*}
  \hat{\y} = \sigma_l(\vW_l \dots\sigma_2(\vW_2\sigma_1(\vW_1\x))).
\end{align*}\end{linenomath*}
The values of the weights are optimized to minimize the squared loss between the predicted value and the target using error backpropagation~\cite{hecht1992theory}.
Since the loss function is highly non-convex, the optimization procedure only finds a local minimum, in contrast to the globally optimal solution found by LRR and KRR, though this is usually irrelevant to the performance of FFNNs~\cite{choromanska2015loss} (see supplementary Section~\ref{ssec:ffnn} for more details).

The FFNN has a series of hyperparameters that need to be optimized, ranging from hyperparameters about the model architecture itself (e.g.\ size and number of layers), to hyperparameters relating
to the training procedure, where we used stochastic gradient descent (SGD) with Nesterov momentum~\cite{bengio2013advances}.

Due to the larger number of hyperparameters to optimize, we selected the best hyperparameters by performing a grid search on a small number of values for each hyperparameter.
The optimal hyperparameters were determined based on the performance on a validation set consisting of a random selection of 15\% of the cycles in the training set. The number of training epochs was chosen using early stopping, with training being stopped if the validation error had not improved in the last six epochs in the case of the synthetic dataset, while a stopping criterion of 30 epochs was used for the real-world dataset due to the overall smaller size of the training set.

As with the other stateless models, the FFNN was trained using inputs from a 24-hour time window into the past for the synthetic dataset, and inputs only at the current time point for the real-world dataset.

\subsection{Echo state networks (ESN)}

ESNs are an alternative RNN architecture that can alleviate some of the training related problems of RNNs (see Section~\ref{ssec:stateful} in the supplement)
by not using error backpropagation for training at all~\cite{jaeger2004harnessing}. Instead, ESNs use very large randomly initialized weight matrices combined with a recurrent mapping of the past inputs; collectively called the ``reservoir''. This way, ESNs can keep track of the hidden state $\h(t) \in \mathbb{R}^{m}$ (with $m >> d_x$) of the system by updating $\h(t)$ as a combination of the previous hidden state $\h(t-1)$ and the current input vector $\x(t)$.
The final prediction of the output is then computed using LRR on the inputs and hidden state, i.e.,
\begin{linenomath*}\begin{align*}
\hat{\y}(t) = \vW_\text{out}[\x(t); \h(t)] \quad \text{ with }\vW_\text{out} \in \mathbb{R}^{d_y \times (d_x + m)}.
\end{align*}\end{linenomath*}
In general, echo state networks are a very powerful type of RNN, whose performance on dynamical system forecasting is often on par with or even better than that of other, more popular and complex RNN models (LSTM, GRU, etc.)~\cite{jaeger2004harnessing,bianchi2017overview}. Since the only learned parameters are the weights $\vW_\text{out}$ of the linear model used for the final prediction, ESNs can also be trained on smaller datasets without risking too much overfitting.

For the ESN model, which had the largest number of hyperparameters (see supplementary Sections~\ref{ssec:esn} and~\ref{ssec:hyperparams}), we used a random search to select candidate hyperparameter sets, which were then evaluated using a 10-fold cross-validation on the training set.

As the ESN is a stateful model, capable of encoding the past in its hidden state, the input vector at any time point $t$ only consists of the process conditions at the current time point, i.e.\ $\x(t)$.

\subsection{LSTM networks}

Another very popular architecture for dealing with the vanishing gradients problem of RNNs is the \emph{long short term
memory} (LSTM) architecture, which was developed specifically for this purpose \cite{hochreiter1997long}. LSTMs are
trained using error backpropagation as usual, but avoid the problem of vanishing gradients by using an additional
state vector called the ``cell state'', alongside the usual hidden state, which is updated slowly via special layers called ``gates'', and thus is capable of preserving long-term information.
While the updates of the hidden state $\h(t)$ of an LSTM network are much more complex compared to ESNs, the final prediction is again only a linear transformation of the network's internal hidden state:
\begin{linenomath*}\begin{align*}
\hat{\y}(t) = \vW_\text{o}\h(t) \quad \text{ with }\vW_\text{o} \in \mathbb{R}^{d_y \times m}.
\end{align*}\end{linenomath*}
However, in this case, the parameter values of $\vW_\text{o}$ are optimized together with the other parameters of the LSTM network, instead of using a separate LRR model.

Due to the multiple layers needed to model the gates that regulate the cell state, the LSTM typically requires larger amounts of training data to avoid overfitting. Despite its complexity, however, the stability of the gradients of the LSTM make it very well suited for time series problems with long ranging dependencies~\cite{graves2014towards,sutskever2014sequence,donahue2015long,karpathy2015unreasonable,oord2016pixel}.

The LSTM hyperparameters are very similar to that of the FFNN and were again selected with a grid search on a small number of values.
Analogously to the FFNN, the different hyperparameter sets were evaluated on a validation set consisting of a random selection of 15\% of the cycles in the training set, and the number of training epochs was determined using early stopping, i.e., training stopped when the validation error did not improve for six epochs for the synthetic dataset and 30 epochs for the real-world dataset.

Since the LSTM is also capable of encoding the past into a hidden state, the input vector at any time point $t$ only consists of the process conditions at the current time point, i.e.\ $\x(t)$.

\section{Results}\label{sec:results}
In this section, we report our evaluation of the five different ML models introduced in Section~\ref{sec:ml_methods} using the synthetic and real-world datasets described in Section~\ref{sec:data}. To measure the prediction errors of the ML models, we use the mean squared error (MSE), which, due to the subdivision of our datasets
into cycles, we define slightly differently than usual: Let the dataset $\D$ be composed of $N$ cycles,
and let $\y_i(t)$ denote the KPIs at time point $t \in \{0, \dots, T_i\}$ within the $i$-th cycle, where
$T_i$ is the length of the $i$-th cycle. Then, given the corresponding model predictions $\hat{\y}_i(t)$,
the MSE of a model for the entire dataset is calculated as
\begin{linenomath*}\begin{align}\label{eq:MSE}
  \mathrm{MSE}(\D) = \frac{1}{N}\sum\limits_{i = 1}^{N} \frac{1}{T_i}\sum\limits_{t=0}^{T_i} \left(\y_i(t) - \hat{\y}_i(t)\right)^2.
\end{align}\end{linenomath*}

Since the synthetic and real-world datasets are very different, they were used
to examine different aspects of the models. The synthetic dataset was used to examine how the models
perform in a nearly ideal scenario, where data is freely available and the noise is very low or even non-existent.
On the other hand, the real-world dataset was used to test the robustness of the models, since it contains only a
limited amount of training samples and a relatively high noise level.

\subsection{Synthetic dataset}
The synthetic dataset was randomly split into a training set (1938 cycles consisting of all together 391876 time points) and a test set (215 cycles / 44041 time points).
For all models, the inputs were all scaled to have zero mean and unit variance, while the KPIs were left unscaled.
Only results for conversion as a degradation KPI are discussed; results for selectivity follow the same trends.

\begin{figure}[htb!]
  \centering
  \includegraphics[width=.9\textwidth]{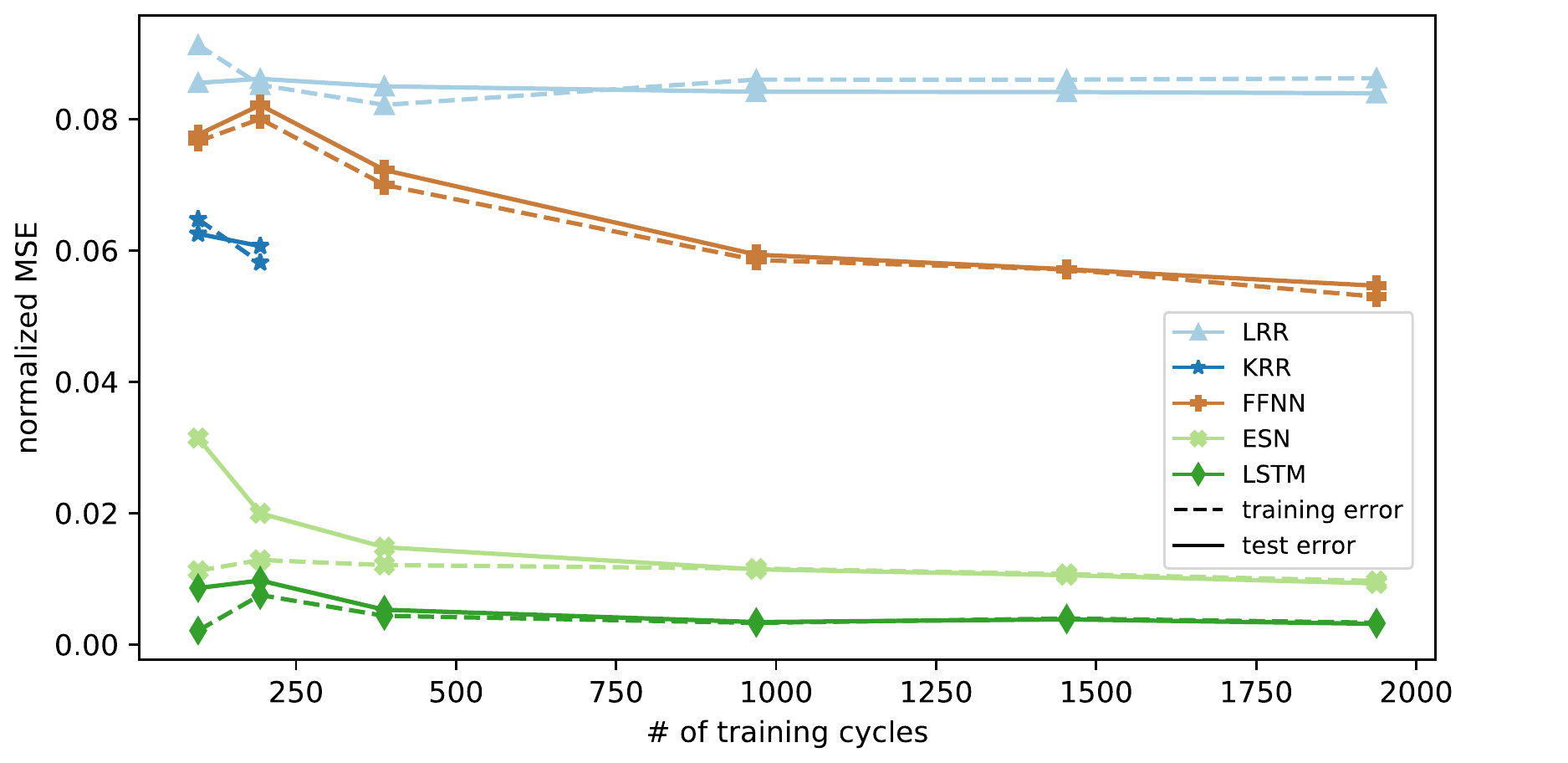}
  \caption{Training and test set MSEs for the five different models, where the models were trained on differently sized subsets of the training set from the synthetic dataset. The performance for KRR cuts off early as it is too computationally expensive to train KRR with the roughly 2000 degradation cycles, corresponding to 400000 time points.}
  \label{fig:var_size_results}
\end{figure}
Fig.~\ref{fig:var_size_results} shows the normalized mean squared errors (MSEs) for each of the models on the training
and test splits when trained on differently sized subsets of the full training set.
For most of the models, the error converges relatively early, meaning that even with a fraction of the complete
dataset, the models manage to learn an accurate approximation of the dynamics of the synthetic dataset, as far as the respective model complexity permits.
This also indicates that the existing errors in the models are largely due to the limitations on the flexibility
of the models themselves, and not due to the training set not being large enough. This is clearly evident
with LRR, which essentially achieves its maximum performance using 5\% of the total dataset size. Since LRR
is a linear model, it can only learn the linear relations between the inputs and outputs. While this high
bias prevents the model from learning most of the non-linear dynamics regardless of the training set size, this also
means that the model has low variance, i.e., it tends not to overfit on the training data~\cite{friedman2001elements}.
For the FFNN, the error slowly declines as the number of samples increases, though at an ever slower rate, with
the error using the full training dataset being significantly lower that LRR.

As for ESN and LSTM, both methods seem to somewhat overfit for the smaller training set sizes, as indicated by the
differences between training and test errors, however, even in the small data regime the test errors are much
lower compared to the three stateless models.
The errors of both stateful models converge at around 50\% of the full dataset, after which there is virtually
no overfitting and no significant improvement of the performance for larger dataset sizes.

The general lack of overfitting can be partially attributed to the lack of noise in the synthetic dataset,
since overfitting usually involves the model fitting the noise instead of the actual signal/patterns.
Additionally, the low amount of overfitting in the small data regime also points to the fact that such time-dependent stateful
models are well-suited for this problem setting.

Across all dataset sizes, the LSTM model is clearly the best performing, with its error when using the full
dataset being 5 times smaller than the error of the ESN model. This trend is also confirmed by other performance metrics (Table~\ref{tab:syn_res}). Additionally, the scale invariant
measures (normalized MSE and $R^2$) show that all of the models have relatively small errors compared to the variance of the
dataset.

Given the great performance of the ESN and especially the LSTM model, these experiment clearly demonstrate
that even with smaller amounts of high-quality data, entire degradation cycles can in principle be predicted with very high accuracy.
\begin{table}[htbp!]
  \centering
    \caption{Regression performance metrics for the different models on the synthetic dataset. $^*$Note that the KRR model was trained using only 10\% of the full training data.}
    \label{tab:syn_res}
    \begin{tabular}{ccccccccc}
      \textbf{Metric} & \multicolumn{2}{c}{\textbf{MSE}} & \multicolumn{2}{c}{\textbf{Norm. MSE}} & \multicolumn{2}{c}{\textbf{MAE}} & \multicolumn{2}{c}{$\mathbf{R^2}$}\\
      \hline
      & train & test & train & test & train & test & train & test \\
      \hline
      LRR     & $2.345$ & $2.307$ & $0.086$ & $0.084$ & $1.002$ & $1.014$ & $0.914$ & $0.916$ \\
      KRR$^*$ & $1.462$ & $1.635$ & $0.053$ & $0.059$ & $0.671$ & $0.745$ & $0.947$ & $0.941$ \\
      FFNN    & $1.433$ & $1.496$ & $0.052$ & $0.054$ & $0.653$ & $0.672$ & $0.948$ & $0.946$ \\
      ESN     & $0.265$ & $0.259$ & $0.009$ & $0.009$ & $0.172$ & $0.173$ & $0.991$ & $0.991$ \\
      LSTM    & $0.083$ & $0.086$ & $0.003$ & $0.003$ & $0.092$ & $0.097$ & $0.997$ & $0.997$ \\
      \hline
    \end{tabular}
\end{table}

\begin{figure}[htbp!]
  \centering
  \begin{subfigure}[b]{0.35\textwidth}
    \centering
    \includegraphics[height=0.22\textheight]{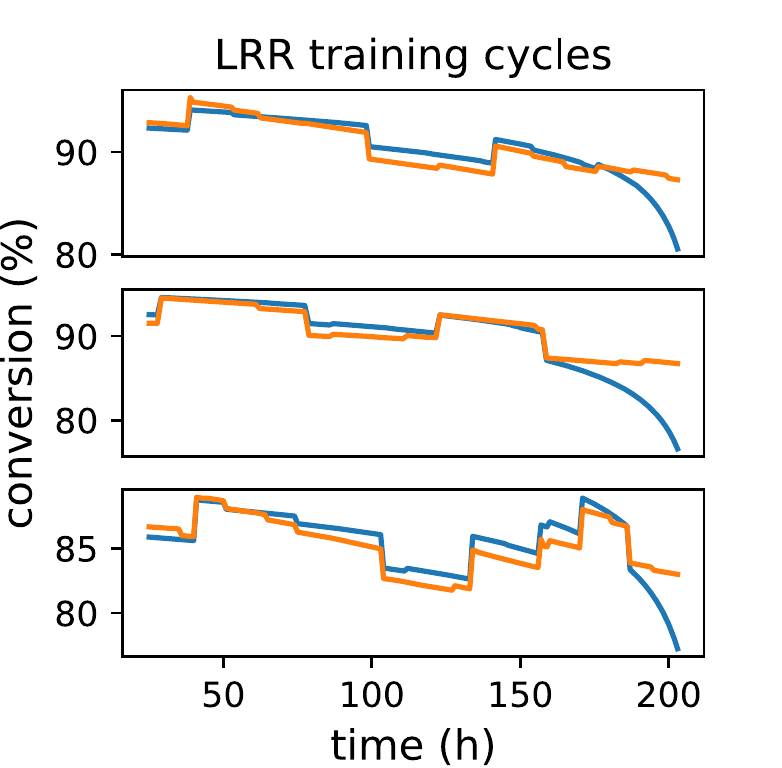}
  \end{subfigure}
  \begin{subfigure}[b]{0.35\textwidth}
    \centering
    \includegraphics[height=0.22\textheight]{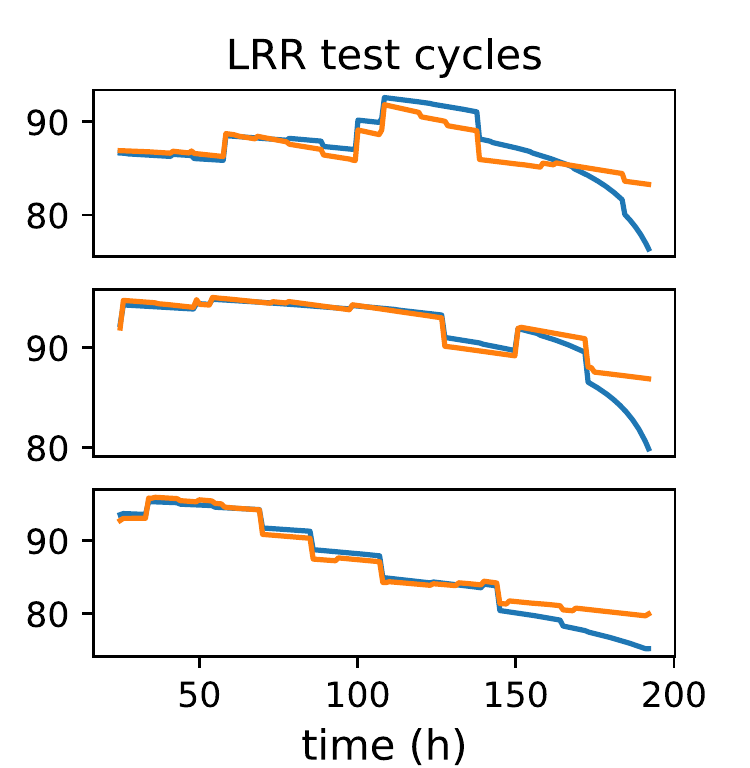}
  \end{subfigure}\\
  \begin{subfigure}[b]{0.35\textwidth}
    \centering
    \includegraphics[height=0.22\textheight]{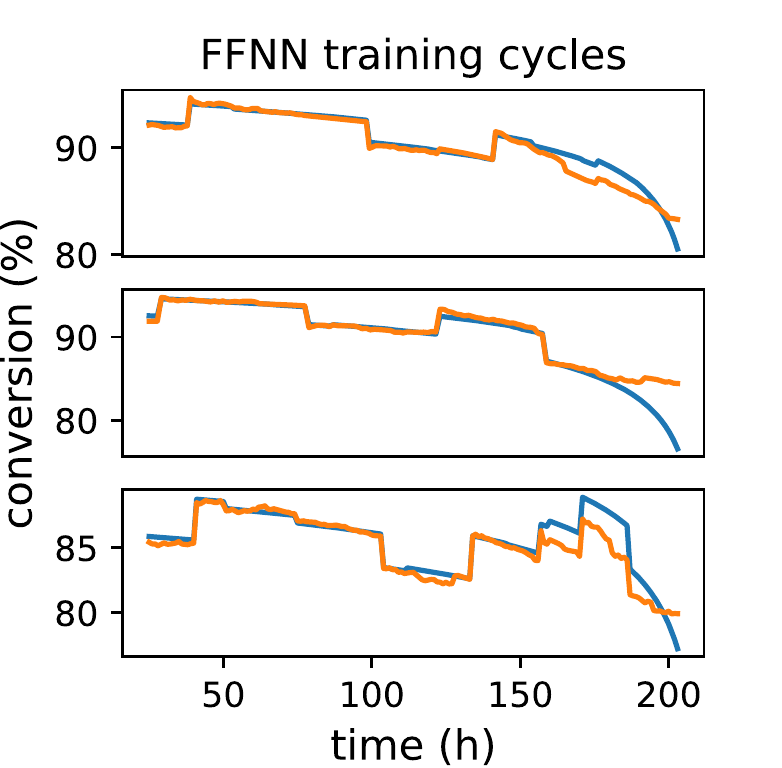}
  \end{subfigure}
  \begin{subfigure}[b]{0.35\textwidth}
    \centering
    \includegraphics[height=0.22\textheight]{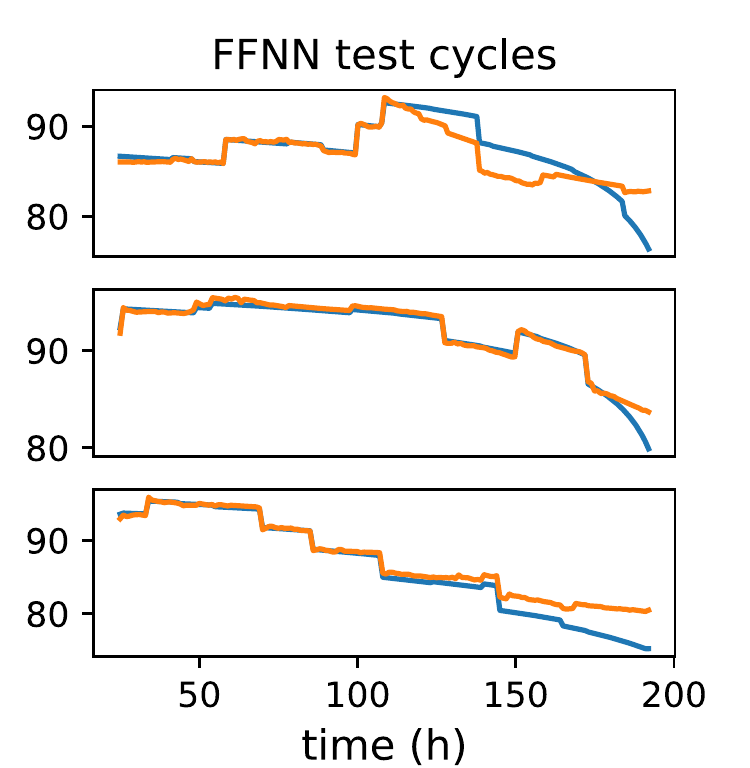}
  \end{subfigure}\\
  \begin{subfigure}[b]{0.35\textwidth}
    \centering
    \includegraphics[height=0.22\textheight]{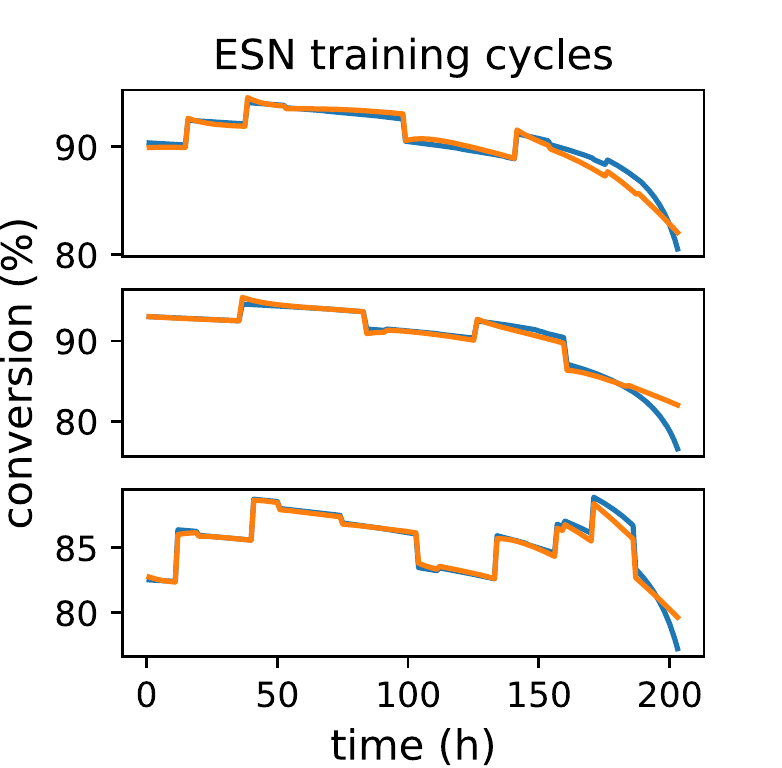}
  \end{subfigure}
  \begin{subfigure}[b]{0.35\textwidth}
    \centering
    \includegraphics[height=0.22\textheight]{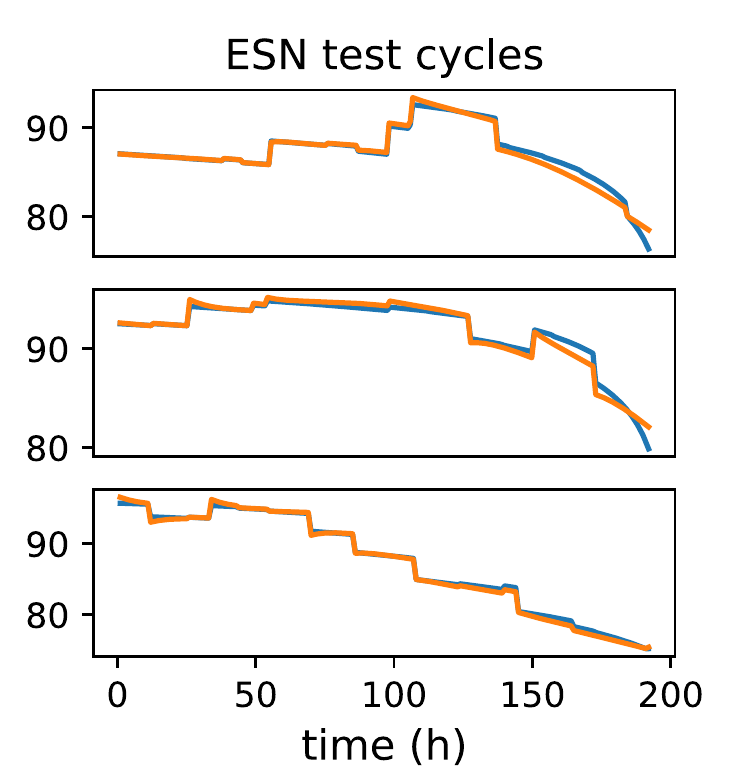}
  \end{subfigure}\\
  \begin{subfigure}[b]{0.35\textwidth}
    \centering
    \includegraphics[height=0.22\textheight]{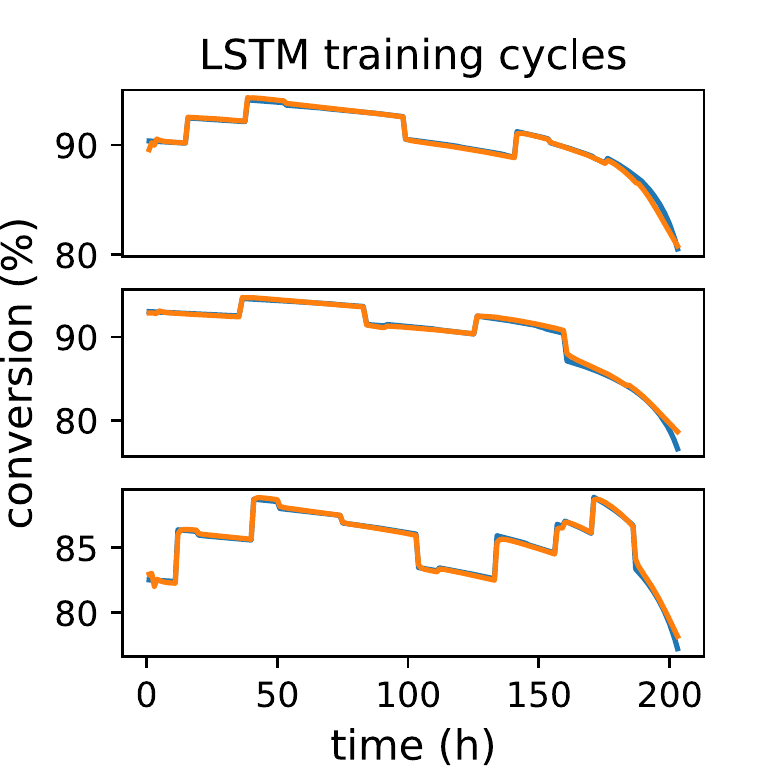}
  \end{subfigure}
  \begin{subfigure}[b]{0.35\textwidth}
    \centering
    \includegraphics[height=0.22\textheight]{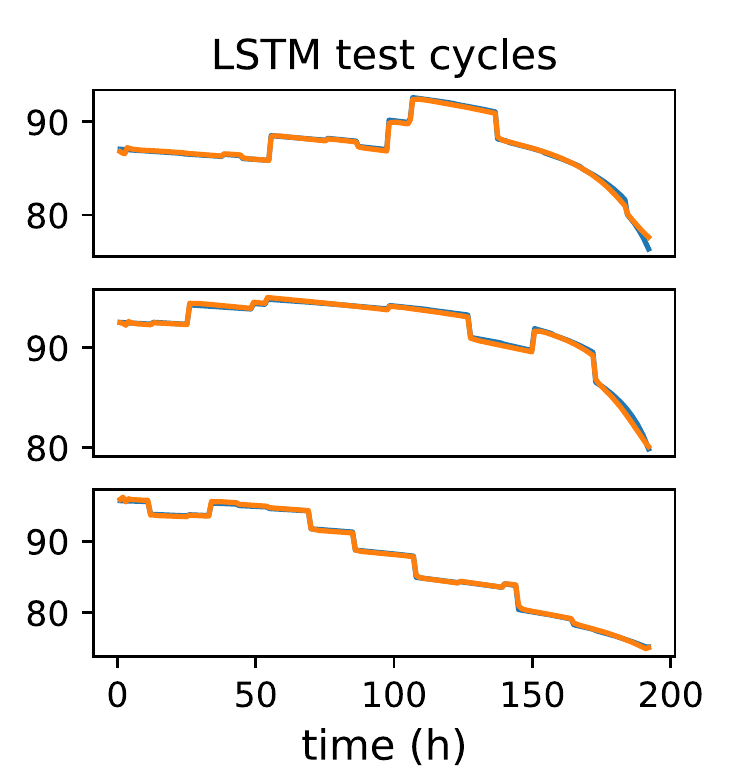}
  \end{subfigure}\\
  \vspace{10pt}
  \begin{subfigure}[b]{\textwidth}
    \centering
    \includegraphics[height=0.025\textheight]{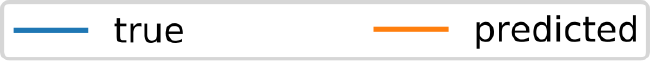}
  \end{subfigure}
  \caption{Comparison of the true and predicted conversion rates across four ML models for some random training and test cycles from the synthetic dataset.}
  \label{fig:synthetic_curves}
\end{figure}
\begin{figure}[htb!]
  \centering
  \includegraphics[width=.9\textwidth]{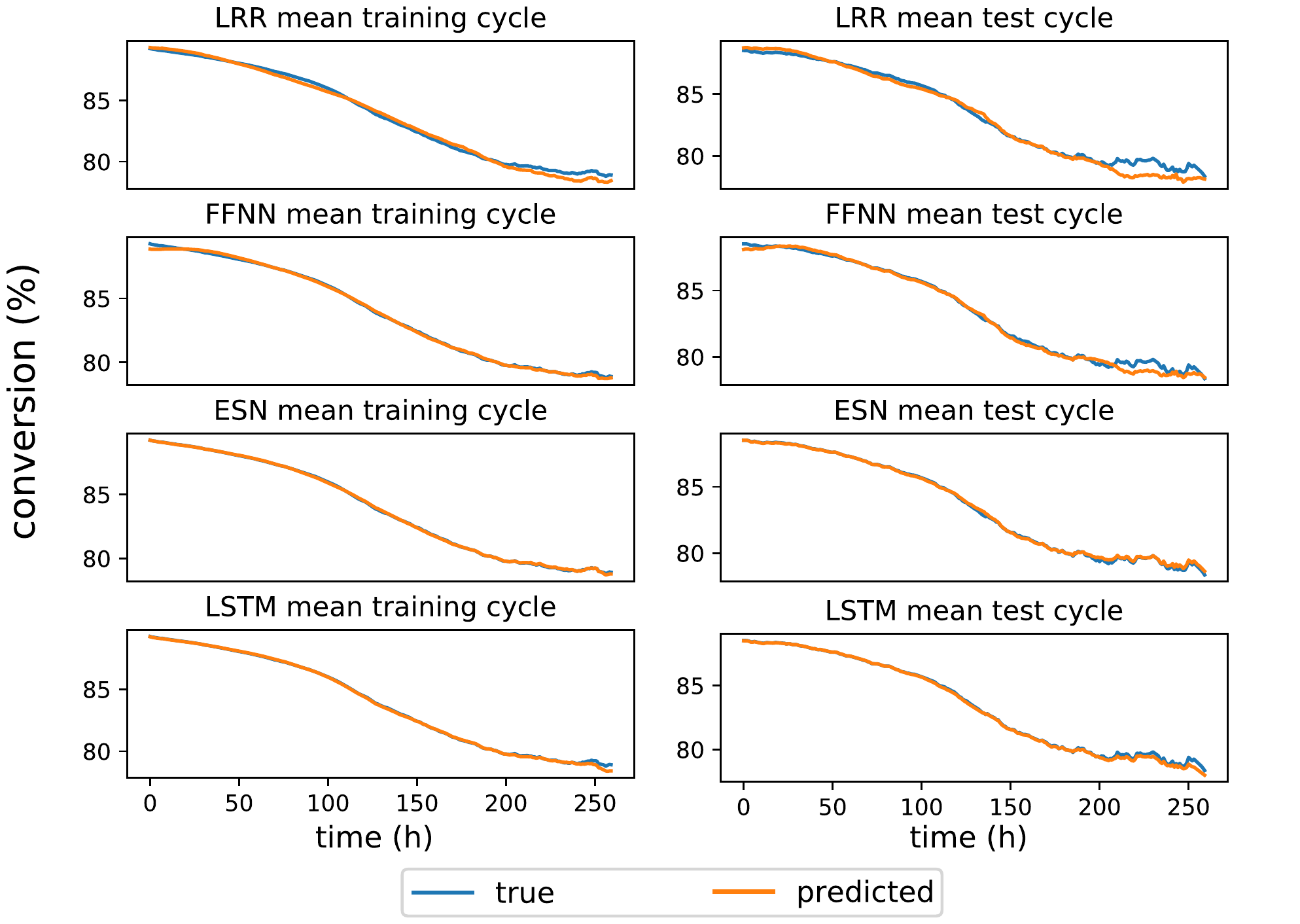}
  \caption{Comparison of the means of the true and predicted conversion rates across four ML models for the training and test cycles of the synthetic dataset.}
  \label{fig:mean_synthetic_curves}
\end{figure}

Fig.~\ref{fig:synthetic_curves} shows plots of the true and predicted conversion rates of the different models
for some randomly selected cycles from the training and test sets. These show that all the models
are capable of accurately predicting the instantaneous effects of the input parameters on the output, since
this relation is largely linear and not time dependent. However, where the models differ the most is in the non-linear
long term degradation, where the stateless models only predict a roughly linear trend, with FFNN coming slightly
closer to the actual degradation trend due to its non-linearity, while the ESN model predicts the
degradation better but fails to capture the rapid decline near the end of each cycle. The LSTM model, on the other hand, manages
to capture the short and long term effects almost perfectly, with only small errors at the very ends of the cycles
where there is smaller amounts of data, due to the varying length of the cycles. This trend is also noticeable on the average
true and predicted KPIs in Figure~\ref{fig:mean_synthetic_curves}, where the LRR and FFNN models have larger differences
especially in the last part of the curves, where the memory effects would be more pronounced, while the mean of the ESN and LSTM predictions have almost perfect alignment with the mean KPI.

\subsection{Real-world dataset}
The real-world dataset is much smaller than the synthetic, consisting of a total of 327 cycles (after outlier removal,
see Section~\ref{sec:data}) with all together 36058 time points. As the real-world dataset stretches over 3 time periods
with different catalyst charges in the reactor, we test the performance in a realistic manner
by selecting the third catalyst charge as the test set, which makes it possible to see to what extent
the models are able to extrapolate across the different conditions caused by the catalyst exchange. This resulted in a
training set that is more than 10 times smaller than the training set of the synthetic dataset, consisting of 256 cycles
(28503 time points), while the test set consists of 71 cycles (7555 time points).
Analogously as with the synthetic dataset, the input variables were all scaled to have zero mean and unit variance, while the KPIs were left unscaled.

\begin{figure}[htb!]
  \centering
  \includegraphics[width=.9\textwidth]{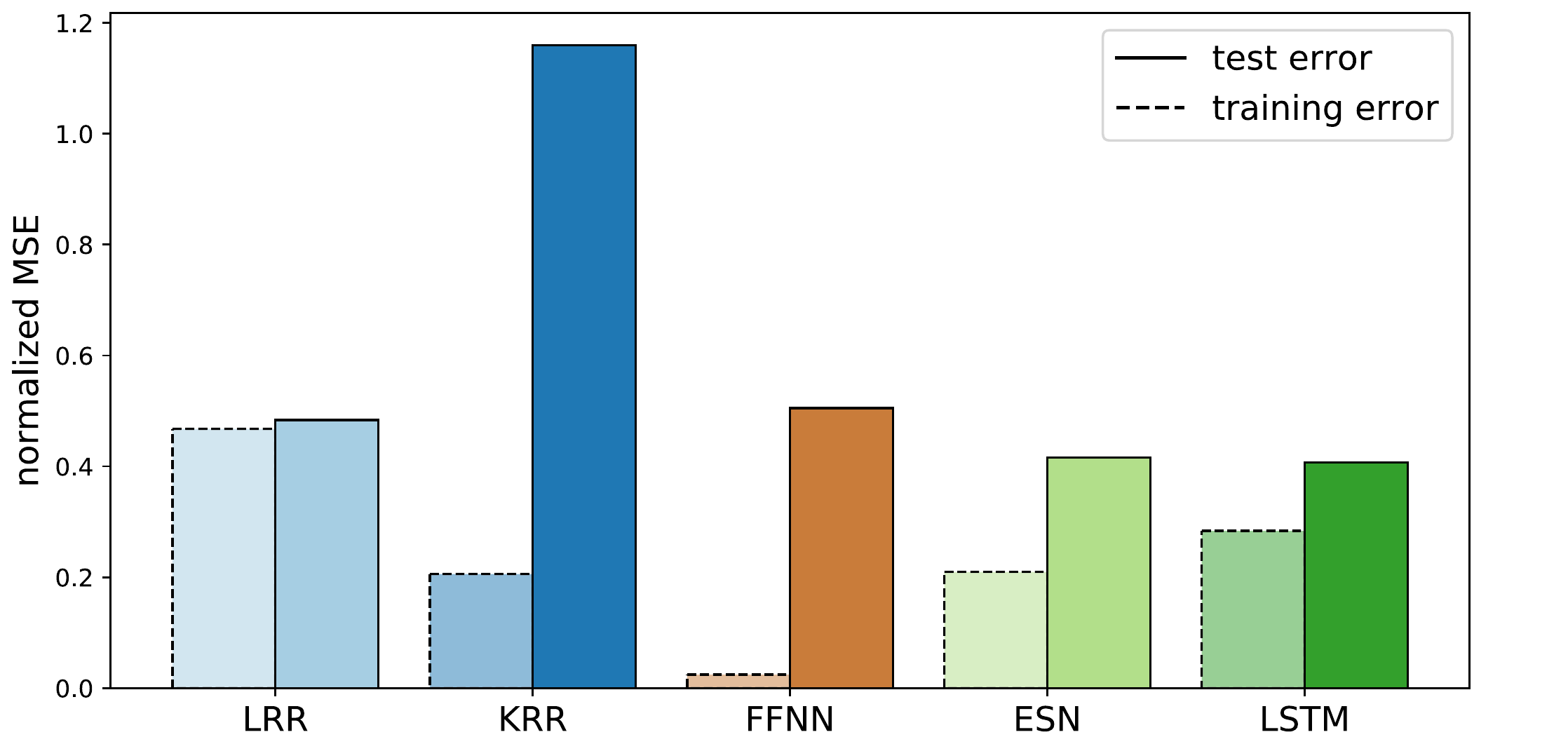}
  \caption{Training and test set MSEs for the five different models evaluated on the real-world dataset.}
  \label{fig:real_world_results}
\end{figure}

Fig.~\ref{fig:real_world_results} shows the normalized mean squared errors for each of the models on the training and test sets.
Due to the larger amounts noise and smaller dataset size, as well as the subtle differences between the training and test distributions due to the different catalyst beds, the results here are different compared to the
synthetic dataset: The more complex models show more overfitting, indicated by the test errors being
significantly larger than the corresponding training errors, especially for KRR, which also has the largest test error of all models.
On the other hand, LRR shows almost no overfit and its performance on the test set is much closer to that of the other models. Once again, ESNs and LSTMs outperform the stateless models, but this time the margin is slimmer and both models show a very similar performance.
The values for multiple other performance metrics (Table~\ref{tab:rw_res}) confirm this conclusion, with the minor differences being found in the mean absolute errors of the model, where the difference in performance between the recurrent models and the rest is more clearly pronounced, and where now the ESN has a slightly better, but still very comparable performance to the LSTM model.
\begin{table}[htbp!]
  \centering
    \caption{Regression performance metrics for the different models on the real-world dataset.}
    \label{tab:rw_res}
    \begin{tabular}{ccccccccc}
      \textbf{Metric} & \multicolumn{2}{c}{\textbf{MSE}} & \multicolumn{2}{c}{\textbf{Norm. MSE}} & \multicolumn{2}{c}{\textbf{MAE}} & \multicolumn{2}{c}{$\mathbf{R^2}$}\\
      \hline
      & train & test & train & test & train & test & train & test\\
      \hline
      LRR     & $36.499$ & $39.620$ & $0.467$ & $0.483$ & $4.350$ & $4.713$ & $0.533$ & $0.517$ \\
      KRR & $16.062$ & $95.053$ & $0.206$ & $1.160$ & $2.398$ & $8.821$ & $0.794$ & $-0.160$ \\
      FFNN    & $1.897$  & $41.386$ & $0.024$ & $0.505$ & $1.442$ & $4.513$ & $0.976$ & $0.495$ \\
      ESN     & $16.364$ & $34.089$ & $0.210$ & $0.416$ & $2.466$ & $2.804$ & $0.790$ & $0.584$ \\
      LSTM    & $22.127$ & $33.353$ & $0.283$ & $0.407$ & $2.632$ & $2.817$ & $0.717$ & $0.593$ \\
      \hline
    \end{tabular}
\end{table}

\begin{figure}[htbp!]
  \centering
  \begin{subfigure}[b]{0.35\textwidth}
    \centering
    \includegraphics[height=0.22\textheight]{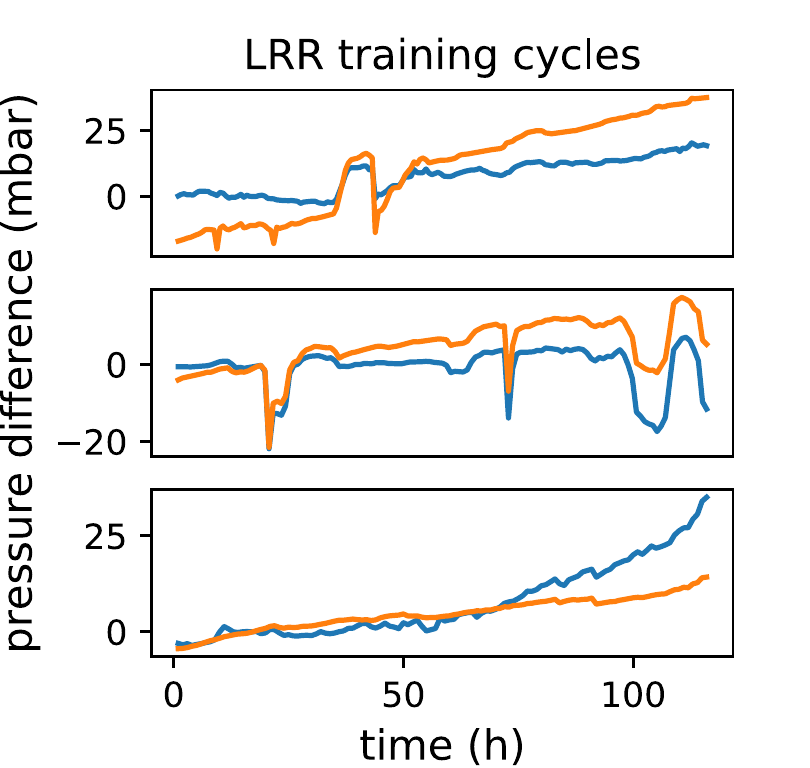}
  \end{subfigure}
  \begin{subfigure}[b]{0.35\textwidth}
    \centering
    \includegraphics[height=0.22\textheight]{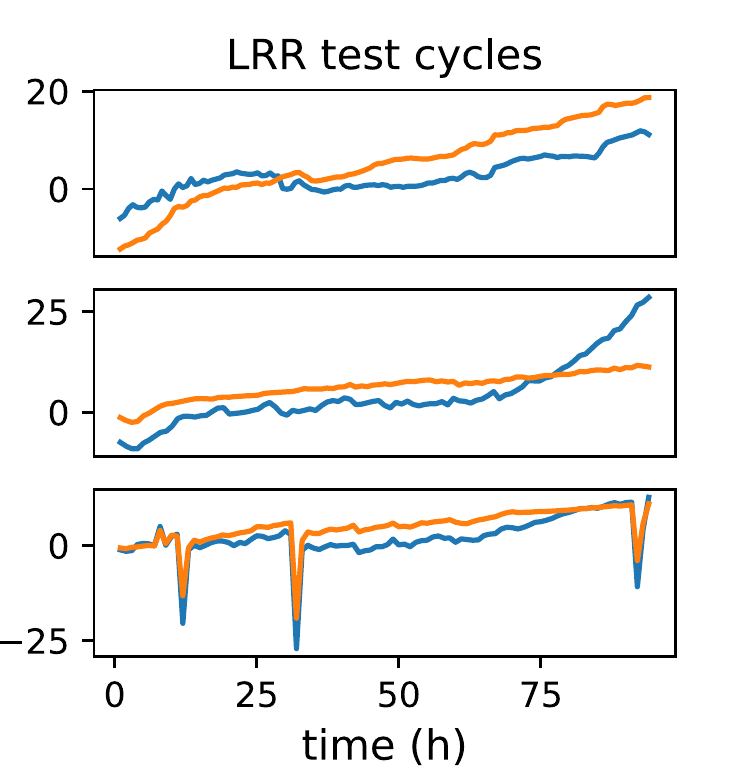}
  \end{subfigure}\\
  \begin{subfigure}[b]{0.35\textwidth}
    \centering
    \includegraphics[height=0.22\textheight]{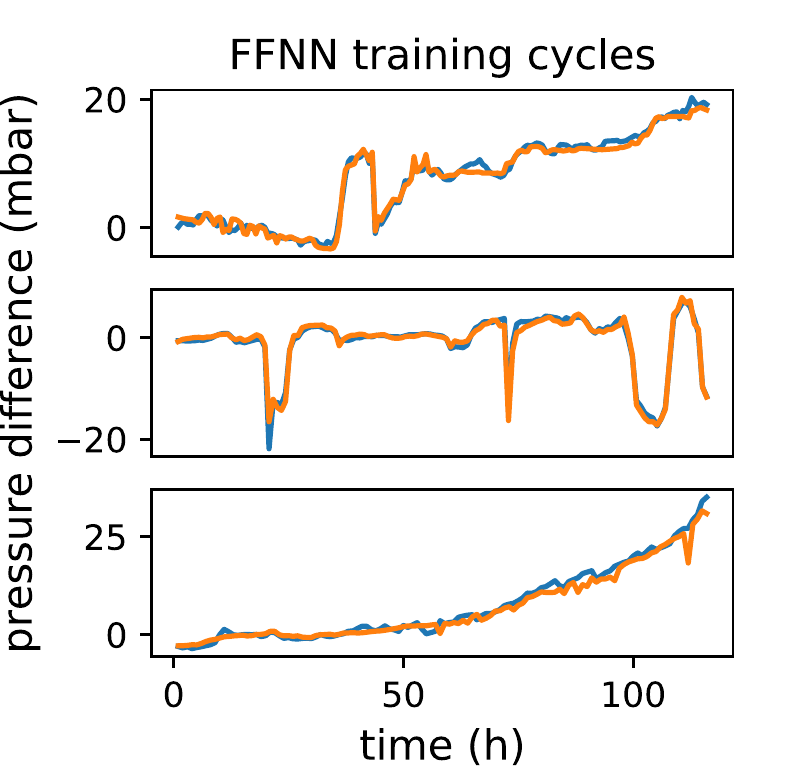}
  \end{subfigure}
  \begin{subfigure}[b]{0.35\textwidth}
    \centering
    \includegraphics[height=0.22\textheight]{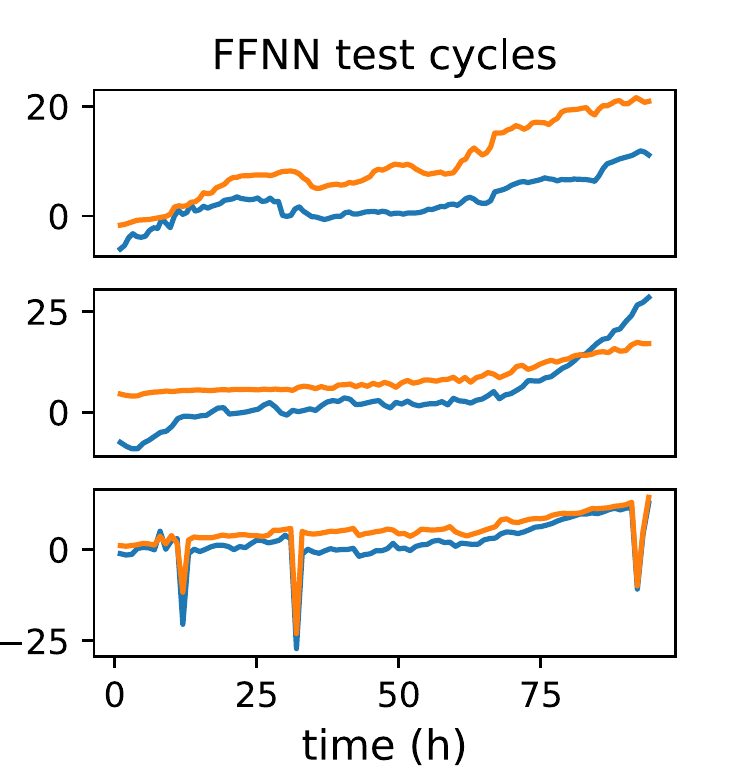}
  \end{subfigure}\\
  \begin{subfigure}[b]{0.35\textwidth}
    \centering
    \includegraphics[height=0.22\textheight]{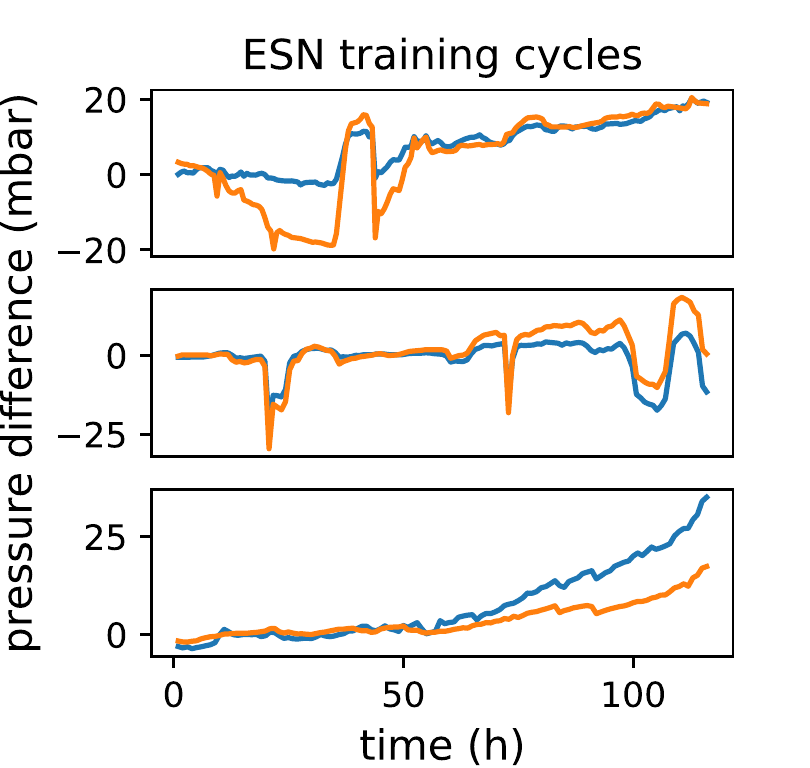}
  \end{subfigure}
  \begin{subfigure}[b]{0.35\textwidth}
    \centering
    \includegraphics[height=0.22\textheight]{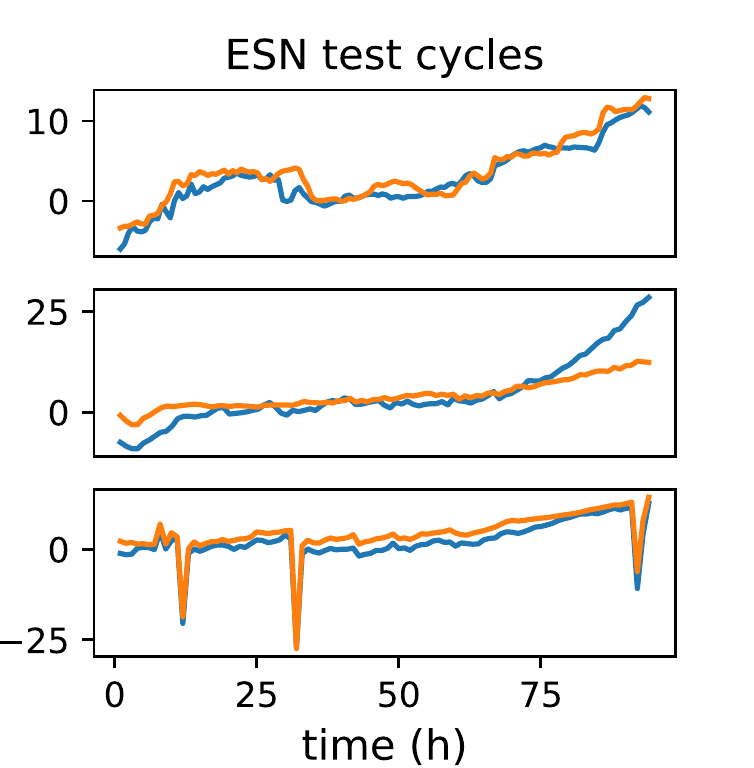}
  \end{subfigure}\\
  \begin{subfigure}[b]{0.35\textwidth}
    \centering
    \includegraphics[height=0.22\textheight]{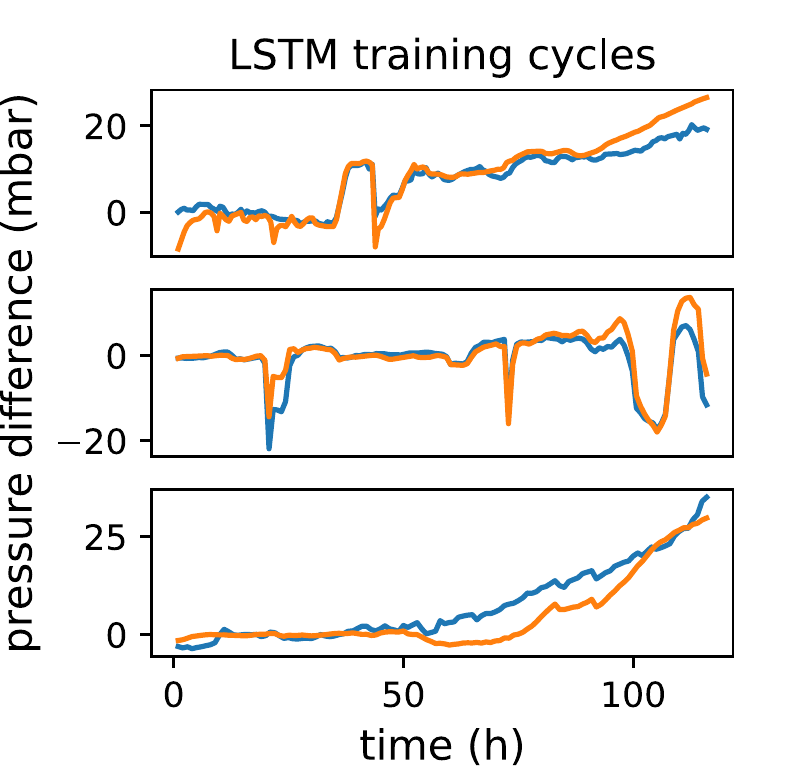}
  \end{subfigure}
  \begin{subfigure}[b]{0.35\textwidth}
    \centering
    \includegraphics[height=0.22\textheight]{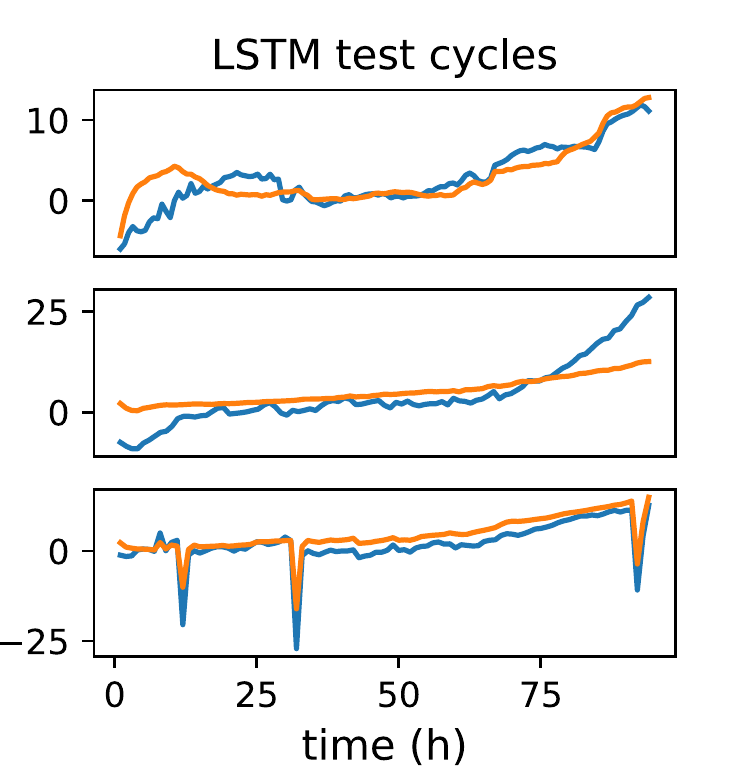}
  \end{subfigure}\\
  \vspace{10pt}
  \begin{subfigure}[b]{\textwidth}
    \centering
    \includegraphics[height=0.025\textheight]{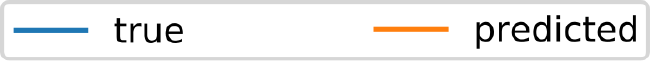}
  \end{subfigure}
  \caption{Comparison of the true and predicted pressure differences across four ML models for some random training and test cycles from the real-world dataset.}
  \label{fig:real_world_curves}
\end{figure}
\begin{figure}[htb!]
  \centering
  \includegraphics[width=.9\textwidth]{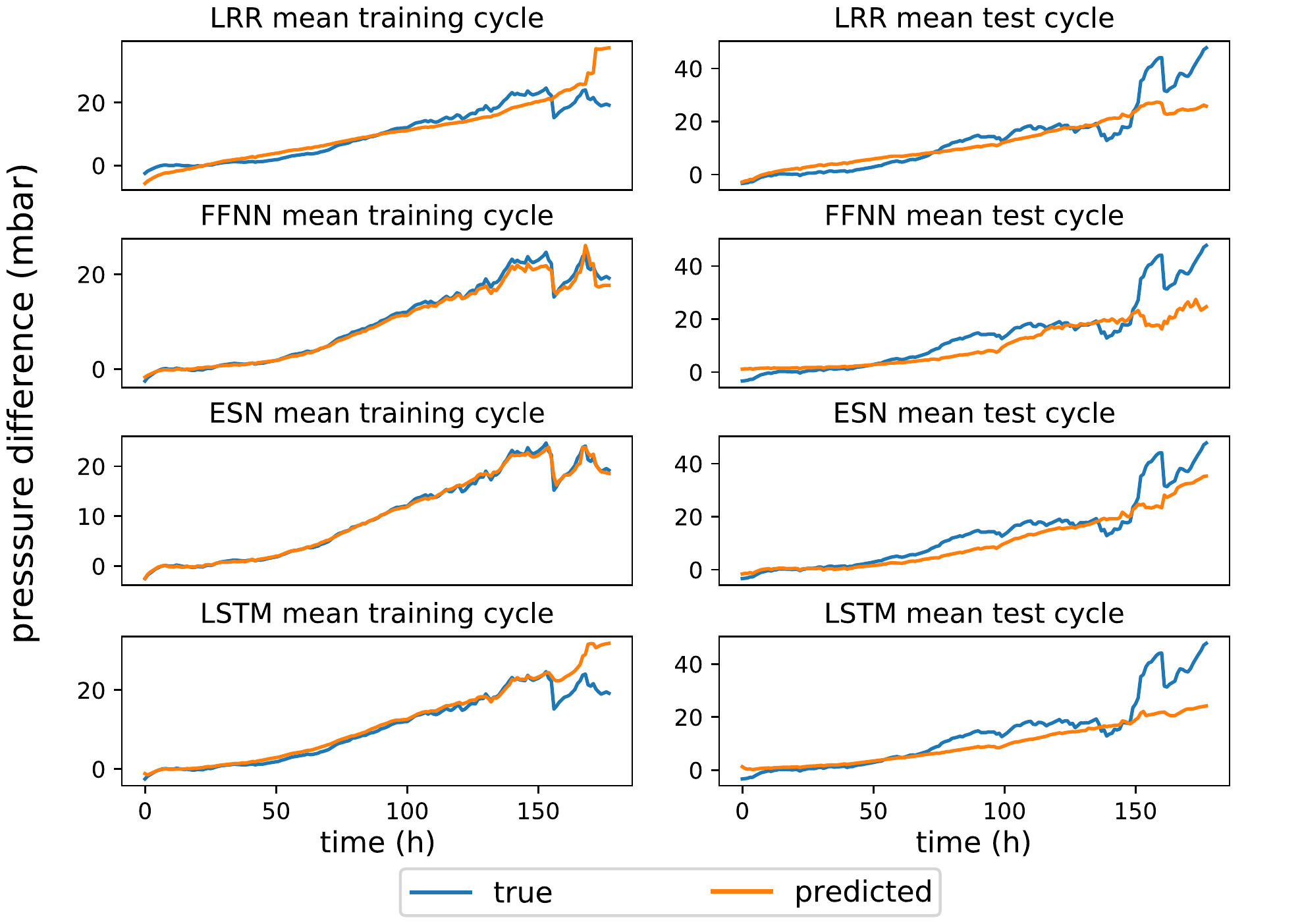}
  \caption{Comparison of the means of the true and predicted pressure differences across four ML models for the training and test cycles of the synthetic dataset.}
  \label{fig:mean_real_world_curves}
\end{figure}

Fig.~\ref{fig:real_world_curves} shows the plots of the true and predicted pressure differences of
the different models for some randomly selected cycles from the training and test sets of the real-world dataset. In this
case, the outputs are much noisier and none of the models captures the dynamics perfectly. Once again all of
the models capture the instantaneous dynamics fairly accurately, though not as well as in the synthetic dataset, and
all of the models struggle with the non-linear degradation trend. The ESN and LSTM models capture
the dynamics in the training set fairly accurately, but due to the differences in dynamics between the training
and test sets, the accuracy of both stateful models is somewhat lower when predicting the degradation trend at the
end of the cycles for the selected test cycles. The average predicted and true KPIs (Fig.~\ref{fig:mean_real_world_curves}) reveal that both LRR and LSTM seem to overestimate the trend at the end of the training cycles on average, while they underestimate it in the test cycles, which is likely a consequence of the covariate shift between the training and test sets of the real-world dataset. All other models also underestimate the exponential trend at the end of the test cycles on average, with the ESN predictions following the trend of the true KPIs most closely. This can point to better a generalization of the ESN in the case of longer sequences.

Overall, these results confirm that more complex stateful models still perform well in a small data regime, even in the
presence of noise and covariate shifts.

\section{Discussion}\label{sec:disc}
Formulating accurate mathematical models of industrial aging processes (IAP) is essential for predicting when critical assets need to be replaced or restored. In world-scale chemical plants such predictions can be of great economic value, as they increase plant reliability and efficiency. While mechanistic models are useful for elucidating the influencing factors of degradation processes under laboratory conditions, it is notoriously difficult to adapt them to the specific circumstances of individual plants. Data-driven machine learning methods, on the other hand, are able to learn a model and make predictions based on the historical data from a specific plant and are therefore capable of adapting effortlessly to a multitude of conditions, provided enough data is available. While simpler, especially linear prediction models have previously been studied in the context of predictive maintenance \cite{LEE2018111}, a detailed examination of more recent and complex ML models, such as recurrent neural networks, was missing so far.

In this paper, we address the task of predicting a KPI, which indicates the slow degradation of critical equipment, over the time frame of an entire degradation cycle, based solely on the initial process conditions and how the process will be operated in this period. To this end, we have compared a total of five different prediction models: three stateless models, namely linear ridge regression (LRR), non-linear kernel ridge regression (KRR) and feed-forward neural networks (FFNN), and two recurrent neural network (RNN) based stateful models, echo state networks (ESN) and LSTMs.
To assess the importance of the amount of available historical data on the models' predictions, we have first tested them on a synthetic dataset, which contained essentially unlimited, noise-free data points. In a second step, we examined how well these results translate to real-world data from a large-scale chemical plant at BASF.

While the stateless models (LRR, KRR, and FFNN) accurately captured instantaneous changes in the KPIs resulting from changing process conditions, they were mostly unable to pick up on the underlying trend caused by the slower degradation effects. ESN and LSTMs, on the other hand, were able to additionally correctly predict long-term changes and obtained higher accuracy without requiring larger amounts of training data compared to the stateless models. With more parameters to tune, the stateful models can overfit on specific patterns observed in the training data, however, the errors associated with overfitting are offset by the increased expressive power and the presence of temporal context, which enable these models to more accurately predict temporally dependent dynamics such as the long term degradation trend. In general, all models, especially those based on RNNs, yielded very promising predictions, which are accurate enough to improve scheduling decisions for maintenance events in production plants. The choice of the optimal model in a particular case depends on the amount of available data. For larger datasets, we found that LSTMs can yield almost perfect forecasts over long horizons. In the case of a smaller, noisy dataset, the stateful models still performed better than the simpler stateless models, disproving the common notion that complex models are always `data hungry'. Nevertheless, if only a small number of cycles are available for training (e.g. less than 100) or the data is very noisy, it can be advantageous to resort to a simpler regression model. In these cases, more extensive feature engineering~\cite{horn2019autofeat} or inclusion of domain knowledge~\cite{von2014hybrid,willis2017simultaneous,asprion2019gray,zendehboudi2018applications,glassey2018hybrid} could help to improve the prediction. Overall, ESNs prove to be a reasonable compromise, as they automatically expand the feature space and keep track of the internal hidden state using the randomly parametrized ``reservoir'' and therefore only require training to fit the set of output weights.

We believe that our work can serve as a basis for further research aimed at forecasting industrial aging processes with machine learning models. While the models examined in this manuscript exhibit good performance in the synthetic and real-world scenarios, there is still a wide range of research directions that can improve the performance.

While machine learning models are very good at interpolating, i.e., predicting values for data points from the same distribution as they were trained on, extrapolating beyond the patterns observed in the historical data is much harder and clearly a limitation of most machine learning models. Unfortunately, this is often what is required when dealing with real-world data. For example, on our real-world dataset, the machine learning models struggled to transfer their model of the training data to the test data, which came from a different catalyst charge and thus from a different distribution than the training data. These types of continuous changes and improvements to the setup of a production plant make it very difficult to compile a large enough dataset containing the relevant information for a model to predict future events. These time dependent changes in the data also require extra care as not to overfit on the past, e.g., when selecting hyperparameters for the models on the training set.
Some of these effects might be tackled by explicitly addressing such non-stationarity between training and test set \cite{sugiyama2007covariate, sugiyama2012machine}. It might furthermore be interesting to examine the effects of training on a larger collection of more diverse historical data and using transfer learning to be able to adapt the models to new datasets with smaller amounts of training data \cite{aswolinskiy2017unsupervised, fawaz2018transfer}. This way, more expressive models such as LSTMs could be trained to learn some general patterns on a larger dataset and then fine-tuned on the more recent time points (which should be closer to the current degradation dynamics) to yield more accurate predictions for the future.

Especially when there is reason to believe that the predictions of a machine learning model may not always be perfectly on point, for example, when the model was trained on a very small dataset and predictions need to be made for inputs from a sightly different distribution, a predictive maintenance system could be further improved by providing confidence intervals, which quantify the uncertainty of the model for individual predictions \cite{efthymiou2012predictive}. This could, for example, be achieved by incorporating probabilistic approaches \cite{blei2017variational,kolassa2016evaluating,kasiviswanathan2013quantification,feindt2006neurobayes}. Predicting intervals instead of point estimates may further facilitate planing by making it possible to assess the best and worst case scenarios.

While accurate predictions of IAPs will improve the production process by allowing for longer planing horizons, ensuring an economic and reliable operation of the plant, the ultimate goal is of course to gain a better understanding of and subsequently minimize the degradation effects themselves. While mechanistic and linear models are fairly straightforward to interpret, neural network models have long been shunned for their nontransparent predictions. However, this is changing thanks to novel interpretation techniques such as layer-wise relevance propagation (LRP) \cite{bach2015pixel,montavon2018methods,montavon2017explaining,arras2017explaining,kindermans2018learning,lapuschkin2019unmasking}, which make it possible to visualize the contributions of individual input dimensions to the final prediction. With such a method, the forecasts of RNNs such as LSTMs could be made more transparent, therefore shedding light on the influencing factors and production conditions contributing to the aging process under investigation, which could furthermore be used to help improve the underlying process engineering \cite{zhao2019causal}.

\section{Acknowledgments}
KRM acknowledges partial financial support by the German Ministry for Education and Research (BMBF) under Grants 01IS14013A-E, 01GQ1115 and 01GQ0850; Deutsche Forschungsgemeinschaft (DFG) under Grant Math+, EXC 2046/1, Project ID 390685689 and by the Technology Promotion (IITP) grant funded by the Korea government (No. 2017-0-00451, No. 2017-0-01779). MB acknowledges financial support by BASF under Project ID 10044628.

\bibliography{references}

\appendix
\section*{Appendices}\label{sec:appendix}
\addcontentsline{toc}{section}{Appendices}
\renewcommand{\thesubsection}{\Alph{subsection}}
\renewcommand{\thetable}{\Alph{subsection}.\arabic{table}}
\renewcommand{\thefigure}{\Alph{subsection}.\arabic{figure}}
\renewcommand{\theequation}{\Alph{subsection}.\arabic{figure}}
\counterwithout{figure}{section}
\counterwithin{figure}{subsection}
\counterwithout{table}{section}
\counterwithin{table}{subsection}
\counterwithout{equation}{section}
\counterwithin{equation}{subsection}
\subsection{Mechanistic model to generate the synthetic dataset}
\label{ssub:appdx_synth_model}

The mechanistic process model generating the dataset has the following ingredients:

\begin{itemize}
\item Mass balance equations for all five relevant chemical species (olefinic reactant, oxygen, oxidized product, CO$_2$, water) in the reactor, which is for simplicity modeled as an isothermal plug flow reactor, assuming ideal gas law. The reaction network consists of the main reaction $r_1$ (olefine + O$_2$ $\rightarrow$ product) and one side reaction $r_2$ (combustion of olefine to CO$_2$).
With this, the mass flow $F_i$ of each species $i=1...5$ at the reactor outlet is determined by the volumetric reaction rates $r_j$, stoichiometric matrix $\nu_{ij}$, molar weights $M_i$, reactor volume $V$, and residence time $t_\textrm{res}$ as follows:
\begin{linenomath*}\begin{align}
F_i^\textrm{(out)} =  F_i^\textrm{(in)} +V M_i \sum_{j=1,2}  \nu_{ij} \int_0^{t_\textrm{res} } r_j(t) \, \textrm{d} t.
\label{eq:SynDat_Fout}
\end{align}\end{linenomath*}
The residence time is approximated as
\begin{linenomath*}\begin{align}
t_\textrm{res} = \frac{V}{F} \sum_i c_i^\textrm{(in)} M_i. 
\label{eq:SynDat_tres}
\end{align}\end{linenomath*}
Here, $F=\sum_i F_i\equiv F^\textrm{(in)}$ is the total mass flow through the reactor, which is conserved, and $c_i^\textrm{(in)}$ are the molar concentrations at the reactor inlet.
With known reactor temperature $T$ and pressure $p$,\footnote{Since pressure drop in the reactor is not modeled, the  pressure is assumed constant in the entire reactor.}  mass flow rates $F_i$ can be converted into mass fractions $\mu_i$ and molar concentrations $c_i$:
\begin{linenomath*}\begin{align}
\mu_i &= \frac{F_i}{F}, \\
c_i &= \frac{p}{RT} \frac{\mu_i/M_i}{\sum_k \mu_k/M_k},
\label{eq:SynDat_mu2c}
\end{align}\end{linenomath*}
where $R$ denotes the universal gas constant.

\item A highly non-linear deactivation law for catalyst activity $A(t)$, which depends on the reaction temperature $T$ (in Kelvin), flow rate $F$, inflowing oxygen ${c}_{\textrm{O}_2}^{(in)}$, activity $A(t)$ itself, and kinetic parameters $k_A$ and $E_{A}$:
\begin{linenomath*}\begin{align}\label{eq:SynDat_DegradDyn}
\frac{\textrm{d}A(t)}{\textrm{d}t} = -k_A \cdot \exp{(-E_{A}/R T)} \cdot [{\mu}_{\textrm{O}_2}^{(in)} \cdot F]^3 \cdot [A(t)]^{-5}. 
\end{align}\end{linenomath*}
Catalyst activity is expressed in relative units (0\% to 100\%). As the deactivation can not be observed directly, $A(t)$ is a hidden state variable of the system.

\item Kinetic laws for the reaction rates $r_j$, with kinetic parameters $k_j$ and $E_j$:
\begin{linenomath*}\begin{align*}
r_1(t) &= A(t) \cdot k_1 \cdot \exp{(-E_{1}/RT)} \cdot {c}_{\textrm{Olef.}}(t) \cdot  {c}_{\textrm{O}_2}(t) \\ 
r_2(t) &=   k_2 \cdot \exp{(-E_{2}/RT)}  \cdot  {c}_{\textrm{Olef.}}(t) \cdot \sqrt{  {c}_{\textrm{O}_2}(t) }
\end{align*}\end{linenomath*}
Note that only the main reaction is catalyzed and depends on $A(t)$.
\item The relation for selectivity and conversion of the process:
\begin{linenomath*}\begin{align}
\label{eq:SynDat_SelConv}
S &= \frac{ {c}_{\textrm{Product}}^{(in)} }{ {c}_{\textrm{Olef.}}^{(in)} - {c}_{\textrm{Olef.}}^{(out)} } \\
C &= \frac{ {c}_{\textrm{Olef.}}^{(in)} - {c}_{\textrm{Olef.}}^{(out)} } { {c}_{\textrm{Olef.}}^{(in)} }
\end{align}\end{linenomath*}

\end{itemize}

Parameter values for {$k_j$, $E_j$, $V$} are provided in Table~\ref{tab:params_syndata}.
\begin{table}[htbp!]
  \centering
    \caption{Parameter values for the generation of the synthetic dataset.}
    \label{tab:params_syndata}
  \begin{tabular}{lll}
    Parameter & Value &  Unit\\
    \hline
      $k_1$ & 30000         & $mol/m^3/s \cdot (mol/m^3)^{-2}$ \\
     $E_1$ & 42                & $kJ/mol$ \\
      $k_2$ &15000          & $mol/m^3/s \cdot (mol/m^3)^{-1.5}$ \\
      $E_2$ & 45               & $kJ/mol$ \\
      $k_A$ & 2.7E-10     &$(kg/h)^{-3} h^{-1}$ \\
      $E_A$ & 50               & $kJ/mol$ \\
      $V$     & 4.712E-02  & $m^3$ \\
      \hline
    \end{tabular}
\end{table}

The cumulative feed of olefine for each cycle is calculated as:
\begin{linenomath*}\begin{align*}
m_\textrm{Olef.}(t)=\int_{t_k}^t \mu_\textrm{Olef.}^{(in)}(t) \cdot F(t) \, \textrm{d}t,
\end{align*}\end{linenomath*}
where $t_k$ is the beginning of the current deactivation cycle, i.e., the last catalyst regeneration event with $t_k \le t$.

With the mechanistic model, the time series of the process dynamics are generated in the following way:
\begin{enumerate}
\item Beginning of new deactivation cycle: Set catalyst activity to $A(t)=100\%$.
\item Pick discrete random values for the process parameters $F$, $T$, $p$, $\mu_i^{(in)}$, according to the procedure described below, and calculate the residence time $t_\textrm{res}$, Eq.~\eqref{eq:SynDat_tres}.
\item Set the duration of the next time window of constant process parameters [in hours] by a random integer number between 1 and 24.
\item For each hourly interval in this time window
  \begin{enumerate}
  \item Integrate the reaction kinetics over the residence time $t_\textrm{res}$.
  \item Calculate the mass flow $F_i^{(out)}$ at reactor outlet, Eq.~\eqref{eq:SynDat_Fout}, convert into molar concentrations, Eq.~\eqref{eq:SynDat_mu2c}, and infer selectivity $S$ and conversion $C$ from it, Eq.~\eqref{eq:SynDat_SelConv}.
  \item If $C<75\%$: End of cycle criterion met. Catalyst is regenerated. Begin new cycle (step 1).
  \item Reduce catalyst activity $A(t)$ according to the deactivation dynamics, Eq.~\eqref{eq:SynDat_DegradDyn}.
  \end{enumerate}
\item Go to step 2.
\end{enumerate}

\paragraph*{Generation of random process parameters}
Each of the process parameters $x_1=F$, $x_2=p$, $x_3=T$, and $x_4=\mu_\textrm{Olef.}$ is generated independently through the following random process:
\begin{enumerate}
\item At initial time $t=0$, pick $x_i(t=0)$ uniformly from the range of potential values $[x_i^\textrm{min}, x_i^\textrm{max} ]$ for that parameter. This range is $[3500, 4000]\, \mathrm{kg/h}$ for $F$; $p$ is in $[1.25, 1.45]\,\mathrm{bar}$; $T$ is in $[500, 510]\,^\circ\mathrm{C}$; and $\mu_\textrm{Olef.}$ in $[47.5\%,52.5\%]$.
\item At the time $t_k$ of the $k$-th change of process parameters, pick $x_i(t_k)$ from a Gaussian distribution that is centered at the previous value $x_i(t_{k-1})$ and has a standard deviation of $(x_i^\textrm{max} - x_i^\textrm{min})/10$.
\item After generating the random values in this way, round them them to six equidistant values in $[x_i^\textrm{min}, x_i^\textrm{max} ]$. This way, e.g., the potential values of mass flows $F$ are $(3500 + n\cdot 100)\,\mathrm{kg/h}$, with $n={0,1,...,5}$.
\end{enumerate}
Finally, set $\mu_{\textrm{O}_2}^{(in)}=1-\mu_\textrm{Olef.}^{(in)}$. The remaining mass fractions at the reactor inlet are zero.

\subsection{ML models in detail}\label{ssec:mldetail}

\subsubsection{Stateless models}\label{ssec:stateless}
Stateless models are machine learning models that base their forecast only on the inputs within a fixed time window in the past, i.e., exactly as stated in Eq.~\eqref{eq:problem_setting}.

Stateless models include most typical machine learning regression models, ranging from linear regression models to many types of neural
networks~\cite{draper2014applied,bishop2006pattern}. The stateless regression models explored in this paper
are linear ridge regression (LRR), kernel ridge regression (KRR), and feed-forward neural networks (FFNN).

\paragraph{Linear ridge regression (LRR)}\label{ssec:rr}
The LRR model assumes that the process parameters $\x$
and KPIs $\y$ at a time $t$ are linearly related. For this purpose, a weight matrix $\vW \in \mathbb{R}^{d_y \times d_x}$ is used
to model how each of the process parameters affects each KPI, which can be used to predict the KPIs in $\y(t)$ up to some
noise $\epsilon(t)$ that can not be explained by the model:
\begin{linenomath*}\begin{align*}
  \y(t) = \underbracket{\vW\x(t)}_{\hat{\y}(t)}  + \epsilon(t).
\end{align*}\end{linenomath*}

In order to reduce the influence of outliers on the model and to prevent overfitting, we use a linear ridge regression
model, which imposes a L2-regularization on the weight matrix of the standard linear regression model. The amount of
regularization is controlled using the regularization parameter $\lambda$. Let $\X \in \R^{d_x \times N}$ denote the matrix containing the inputs of all cycles concatenated along the time dimension, while
$\Y \in \R^{d_y \times N}$ denotes the corresponding output matrix constructed analogously to $\X$. The optimal weight
matrix $\vW$ can then be found by solving the following optimization problem:
\begin{linenomath*}\begin{align}\label{eq:rr_opt}
  \min\limits_{\vW^*} \|\vW^{T}\X - \Y\|^2 + \lambda\|\vW\|^2.
\end{align}\end{linenomath*}
The advantage of
ridge regression is that the optimization problem can be solved analytically to obtain the globally optimal weight
matrix $\vW$ as
\begin{linenomath*}\begin{align*}
  \vW^* = (\X\X^{T} +\lambda\I)^{-1}\X^{T}\Y.
\end{align*}\end{linenomath*}

Despite its relative simplicity, LRR is widely used in many application scenarios and can often
be used to approximate real-world processes at fairly high accuracies, especially if additional (non-linear)
(hand-)engineered features are available \cite{horn2019autofeat}. Furthermore, considering
the limited amount of training data that is usually available for real-world IAP problems, reliably estimating the parameters of more complex non-linear prediction models such as deep neural networks
needs to be done with great care~\cite{lecun2012efficient}, while linear models provide a more robust solution as they provide a globally optimal solution and are less likely to overfit given their linear nature.

\paragraph{Kernel ridge regression}\label{ssec:krr}
The linear ridge regression optimization problem in Eq.~\eqref{eq:rr_opt} is rewritten by applying the feature map $\phi$ on the inputs to get
\begin{linenomath*}\begin{align*}
  \min\limits_{\vW^*} \|\vW^{T}\phi(\X) - \Y\|^2 + \lambda\|\vW\|^2,
\end{align*}\end{linenomath*}
leading to the analogous analytic solution
\begin{linenomath*}\begin{align*}
  \vW^* = (\phi(\X)\phi(\X)^{T} +\lambda\I)^{-1}\phi(\X)^{T}\Y.
\end{align*}\end{linenomath*}
Since the feature space is not known explicitly, $\vW$ can not be calculated directly. However, using some algebraic transformations
and the kernel trick~\cite{scholkopf1998nonlinear, muller2001introduction,scholkopf2002learning}, the following solution can be derived:
\begin{linenomath*}\begin{align*}
  \hat{\y}' = {\phi(\x')}^T\vW &= {\phi(\x')}^T{\left(\phi(\X){\phi(\X)}^T + \lambda\I\right)}^{-1}\phi(\X)\Y\\
  &= {\phi(\x')}^T\phi(\X){\left({\phi(\X)}^T\phi(\X) + \lambda\I\right)}^{-1}\Y\\
  &= \kr'\underbrace{{\left(\Kr + \lambda\I\right)}^{-1}\Y}_{\balpha} = \sum\limits_{i=1}^N \alpha_{i}k(\x', \x_{i}).
\end{align*}\end{linenomath*}

The kernel function used in this paper is the radial basis function (RBF) kernel, also known as the Gaussian kernel:
\[k(\x,\x') = \exp\left(-\frac{\lVert \x-\x' \rVert^2}{2\sigma^2}\right).\]

The kernel width parameter $\sigma$, along with the regularization parameter $\lambda$ results in a total of two
hyperparameters to optimize for the KRR model.

\begin{figure}[htbp!]
  \centering
  \includegraphics[width=.7\textwidth]{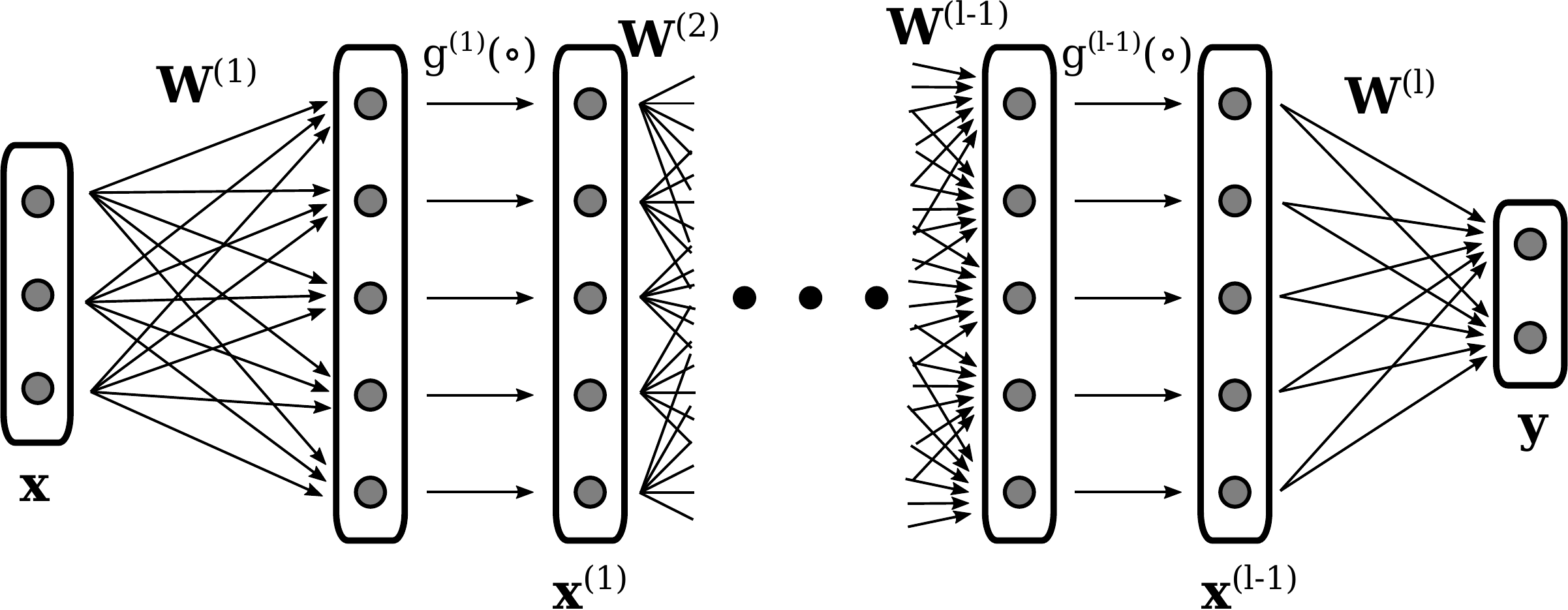}
  \caption{An overview of a basic FFNN architecture with multiple hidden layers with non-linearities.}
  \label{fig:ffn_sketch}
\end{figure}

\paragraph{Feed-forward neural networks (FFNN)}\label{ssec:ffnn}
Let $\x^{(l)}$ denote the feature vector obtained after the transformation at the $l$-th layer, and let $\vW^{(l)}$ be the linear
transformation matrix, $\vb^{(l)}$ the bias and $g^{(l)}$ the non-linear function applied at layer $l$.
Then a generic $l$-layer feed-forward neural network can be defined by the following sequence of operations:
\begin{linenomath*}\begin{align*}
  \x^{(1)} &= g^{(1)}\left(\vW^{(1)}\x + \vb^{(1)}\right)\\
  \x^{(2)} &= g^{(2)}\left(\vW^{(1)}\x^{(1)} + \vb^{(2)}\right)\\
  &\cdots\\
  \x^{(l)} = \hat{\y} &= g^{(l)}\left(\vW^{(l)}\x^{(l-1)} + \vb^{(l)}\right).
\end{align*}\end{linenomath*}
An illustration of this typical FFNN architecture is also shown in Fig.~\ref{fig:ffn_sketch}.

The 1st layer is usually called the input layer and the $l$-th layer the output layer, while all the
layers in-between are known as hidden layers. By changing the values of the weight matrices of the different
layers, the network can be adjusted to approximate the target function. In fact, multi-layer FFNN are proven to be
universal function approximators, meaning that the weights can be adjusted to fit any continuous function~\cite{hornik1989multilayer}.
The most common method for training neural networks to fit a given function is to use error backpropagation to
iteratively update the values of the parameters of the network, usually consisting of the weight matrices and the bias vectors \cite{lecun2012efficient}.
Error backpropagation is a way to efficiently perform gradient descent on the FFNN parameters by minimizing the error/loss
of the network's outputs with respect to some target labels. One of the most commonly used loss functions for regression
problems, and also the one we use in this paper, is the squared error (SE), which for a given FFNN, denoted by $f_{\theta}$, where $\theta$ denotes
the network's parameters, an input time series $\x$, and corresponding
target labels $\y$ is given by:
\begin{linenomath*}\begin{align*}
  \mathds{L}(\x, \y) = (f_{\theta}(\x) - \y)^2.
\end{align*}\end{linenomath*}
By taking the gradient of the loss function with respect to the FFNN's parameters, using backpropagation, one can efficiently perform gradient descent to update the
parameters and minimize the loss of the network \cite{bishop1995neural}. One apparent weakness of gradient descent it that it often converges at local minima, in contrast to
LRR and KRR, where the globally optimal solution to the optimization problem can be obtained analytically. However, in neural networks the losses in these local minima are
often similar to the global optimum~\cite{choromanska2015loss}, so this properties does not significantly impact the performance of a properly trained neural network.
Additionally, due to a FFNN's large number of parameters ($\vW_1, \dots, \vW_l$) and high flexibility, if not properly trained it may overfit, especially when using smaller training sets.

\subsubsection{Stateful models}\label{ssec:stateful}

In contrast to stateless models, stateful models only explicitly use the input $\x(t)$, not the past inputs $\x(t-1),\dots,\x(t-k)$, to forecast the output $\y(t)$
for some time point $t$. Instead, they maintain a hidden state $\h(t)$ of the system that is continuously updated with each new time step and thus
contains information about the entire past of the time series. The output can then be predicted utilizing both the current input conditions, as well as the hidden state of the model: $\hat{\y}(t) = f(\x(t); \h(t))$.

The two stateful models that we are considering for this paper both belong to the class of recurrent neural networks (RNNs).
RNNs are a powerful method for modeling time series, however they can be difficult to
train since their depth increases with the length of the time series. If training is not performed carefully, this can lead to bifurcations of the gradient during the error backpropagation training procedure, which can result in a very slow convergence (``vanishing gradients problem''), if the optimization converges at all \cite{doya1992bifurcations,pascanu2013difficulty}.

The most commonly used stateful models for the modeling of sequential data
are recurrent neural networks (RNNs)~\cite{mandic2001recurrent}. While RNNs are some of the most powerful neural
networks, capable of approximating any function or algorithm~\cite{siegelmann1995computational}, they are also
more involved to train~\cite{doya1992bifurcations,pascanu2013difficulty}.
Consequently, in this paper we chose to model IAPs using two different RNN architectures that are designed precisely to deal
with the problems arising while training regular RNNs: echo state networks (ESN) and long short term memory
(LSTM) networks.

\paragraph{Echo state networks (ESN)}\label{ssec:esn}
\begin{figure}[htbp!]
  \centering
  \includegraphics[width=.7\textwidth]{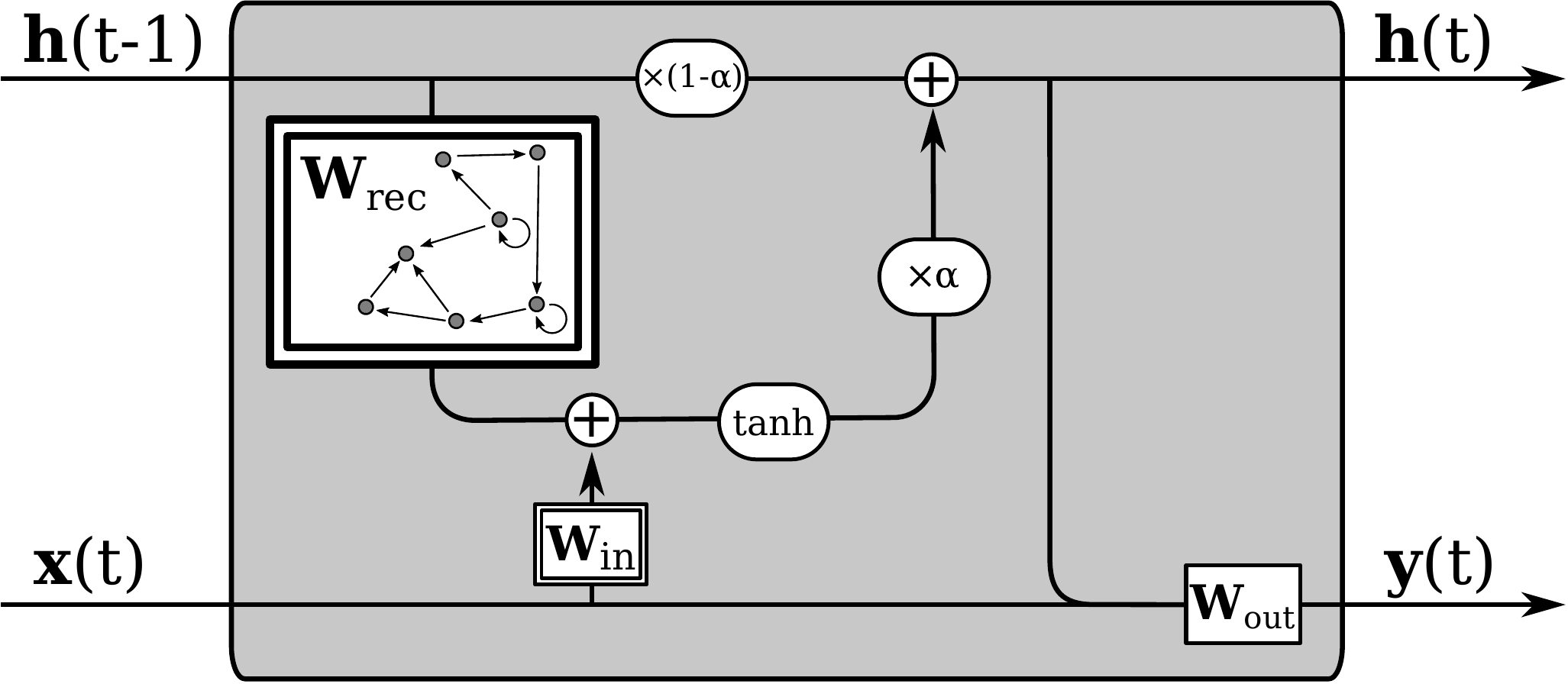}
  \caption{An overview of the ESN architecture.}
\end{figure}
  \label{fig:esn_model}
The basic architecture of an ESN is displayed in Fig.~\ref{fig:esn_model}.
The ESN is composed of two randomly
initialized weight matrices, the input weight matrix $\vW_{in} \in \R^{d_x \times m}$ and the (usually sparse) recurrent
weight matrix $\vW_{rec} \in \R^{m \times m}$, where $m$ is the dimensionality of the hidden state vector and is usually
much larger than the dimensionality of the input features $d_x$. We will refer to these two matrices collectively as
the \emph{reservoir}.
Let $\h(t) \in \R^m$ be the hidden state vector at time $t$ and $\tilde{\h}(t)$ its update, while $\alpha\in(0, 1]$ is
called the leaking rate determining trade-off between adding new and keeping old information in the hidden state.
Additionally, let $[\cdot;\cdot]$ denote vertical vector/matrix concatenation. Now given an input vector $\x(t)$,
the update for the hidden state is given by
\begin{align}\label{eq:esn_update}
  \tilde{\h}(t) &= tanh(\vW_{in}[1;\x(t)] + \vW_{rec}\h(t-1))\\
  \h(t) &= (1-\alpha)\h(t-1) + \alpha\tilde{\h}(t).\nonumber
\end{align}
Note that $tanh$ is applied element-wise here.
It may be counter-intuitive that the randomly generated matrices $\vW_{in}$ and $\vW_{rec}$ can produce
useful features without being trained, but because $m>>d$ the transformation in Eq.~\eqref{eq:esn_update} can be
interpreted as a random feature map that produces a non-linear, high dimensional feature expansion with memory of
the previous inputs. In this sense, one can draw a parallel to kernel methods, which also map the input features to
a non-linear and high dimensional features space, where the features are more easily linearly
separable~\cite{lukovsevivcius2012practical}.

Finally, the predictions are made based on the time-dependent features generated by Eq.~\eqref{eq:esn_update} using a
final output matrix $\vW_{out} \in \R^{d_y \times (m + d_x + 1)}$ as
  \begin{align*}
    \hat{\y}(t) = \vW_{out}[1;\x(t);\h(t)],
  \end{align*}
where $\vW_{out}$ is trained to minimize the MSE of $\hat{\y}(t)$, in our case
using LRR as described in Section~\ref{ssec:rr}. Using LRR provides a globally optimal analytical solution for
$\vW_{out}$, making training the ESN simple and avoiding the problems that arise when training RNNs with
error backpropagation. Due to the high dimensionality and non-linearity of the hidden state features, a simple
model like LRR can still fit complex and non-linear dependencies between the input and the output features.
Additionally, since LRR is very robust against overfitting, ESNs are also very well suited for problems with
smaller amounts of data.

One downside to ESN is that their performance is strongly dependent on the choice of hyperparameters, which mostly
regulate the initialization of the reservoir. Since the reservoir weights are not trained, this initialization is
crucial in order to ensure that the ESN has desirable properties.
The main hyperparameters of the ESN include the dimensionality of the hidden state $m$, the
sparsity of the matrix $\vW_{rec}$, the distribution from which the non-zero elements of the
reservoir are sampled, the spectral radii of each of the reservoir matrices, the scaling of the input
data and finally the leakage rate $\alpha$. These hyperparameters, as well as general
recommendations for their choice are summarized in Table~\ref{tab:esp_hyp} (for more details see~\cite{lukovsevivcius2012practical}).
\begin{table}[htbp!]
  \centering
  \caption{Summary of the main ESN hyperparameters alongside general recommendations for choosing their values.}
  \label{tab:esp_hyp}
  \begin{tabularx}{\textwidth}{lX}
    Parameter & Recommended choice\\
    \hline
    Hidden state size &  Choose as large as you can afford.\\
    Connectivity of $\vW_{rec}$ & A small fixed number of non-zero elements
    (e.g.\ 10) per row on average, irrespective of the hidden state size. \\
    Distribution of non-zero values & Symmetric distribution centered around 0.\\
    Spectral radius of $\vW_{rec}$ & Smaller than or close to 1.\\
    Spectral radius of $\vW_{in}$ & Larger for highly non-linear and small for linear tasks\\
    Leakage rate & Adjust according to the dynamics of the time signal; if signal is changing slowly set close to zero,
    for fast dynamics set close to 1\\
    \hline
  \end{tabularx}
\end{table}

One of the most important hyperparameters is the spectral radius of the $\vW_{rec}$ matrix, because keeping this close
to 1 helps maintain the \emph{echo state property} of the network, which is essential for the ESN to be well-behaved. In a
nutshell, the echo state property means that the hidden state $\h(t)$ should be uniquely defined by the fading history
of the input $\x$, i.e., for a long enough input $\x(t)$, the hidden state $\h(t)$ should not depend on the initialization of the reservoir anymore.

\paragraph{LSTM networks}\label{ssec:lstm}
\begin{figure}[htbp!]
  \centering
  \includegraphics[width=.7\textwidth]{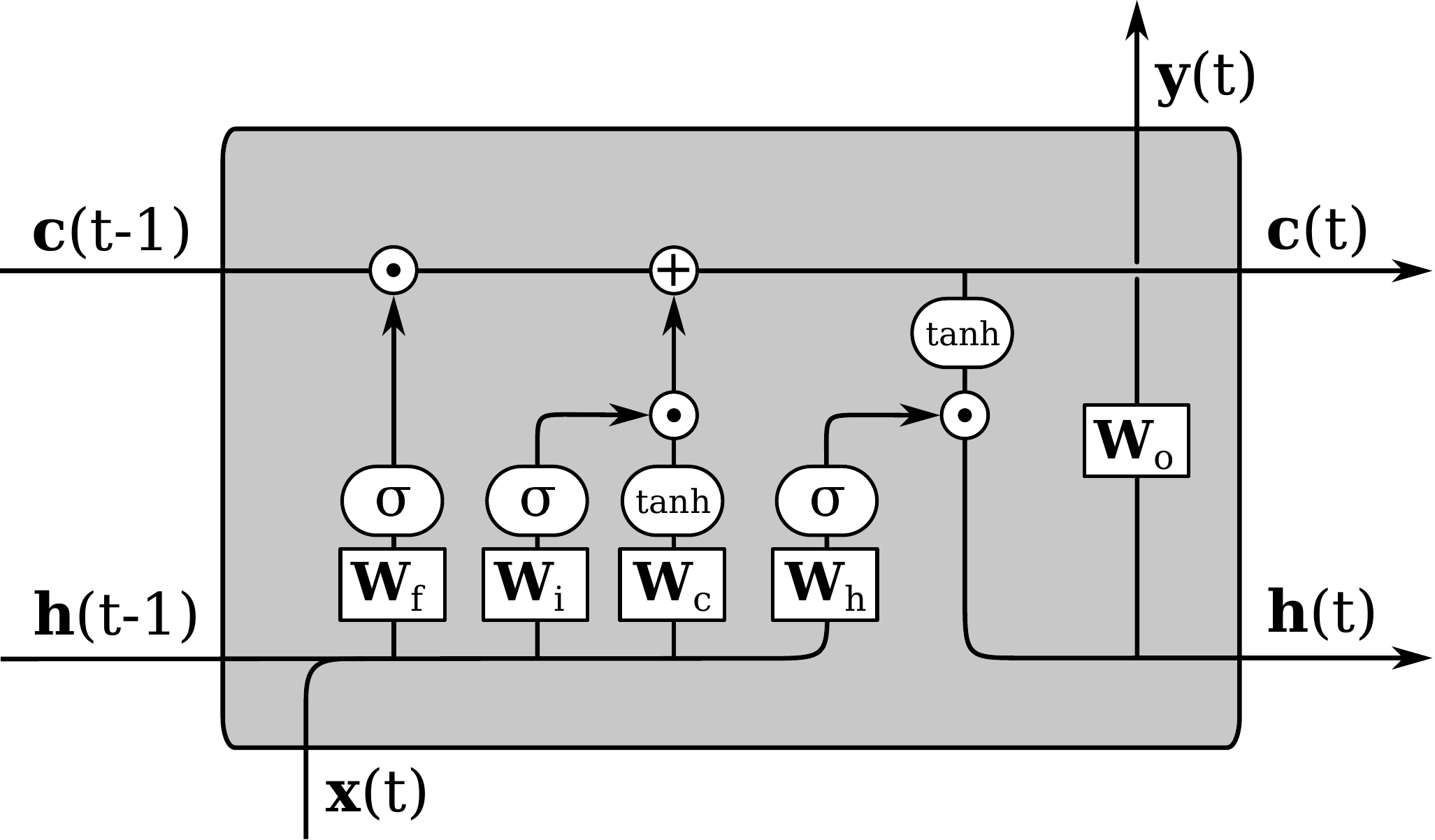}
  \caption{An overview of the LSTM architecture.}
  \label{fig:lstm_model}
\end{figure}
The cell state is a unique and core component of the
LSTM and runs through the entire recurrent chain. It can only be updated slowly at each time step using only linear
updates, making it capable of preserving long term dependencies in the data and maintaining a stable gradient
over long sequences. The inclusion of new or removal of old information to the cell state is carefully regulated
by special neural network layers called gates.
Fig.~\ref{fig:lstm_model} shows the full architecture of one recurrent LSTM cell.
The first gate of the LSTM is the
\emph{forget gate}, which is trained to regulate which information is to be removed or `forgotten' from the cell state.
The forget gate produces a vector $\vf(t)$ that is calculated by applying the sigmoid function, denoted by $\sigma$, element-wise
to a linear transformation of the current input $\x(t)$ and the previous hidden state $\h(t-1)$, resulting in the equation
\begin{linenomath*}\begin{align*}
  \vf(t) = \sigma(\vW_f[\h(t-1), \x(t)] + \vb_f).
\end{align*}\end{linenomath*}
The non-linearity $\sigma$ ensures that the values of $\vf(t)$ are between 0 and 1, with a value of 0 of $\vf_i(t)$ meaning
forget everything at position $i$ in the cell state, and a value of 1 meaning preserve the information at this
position completely. Analogously to the forget gate, the input gate creates a vector $\vi(t)$, which regulates which
values of the cell state will be updated. Additionally, another layer is used to create a vector $\tilde{\vc}(t)$ of
candidate update values that could be added to the cell state based on $\vi(t)$. This results in the following two
equations:
\begin{linenomath*}\begin{align*}
  \vi(t) &= \sigma(\vW_i[\h(t-1), \x(t)] + \vb_i) \\
  \tilde{\vc}(t) &= tanh(\vW_c[\h(t-1), \x(t)] + \vb_c),
\end{align*}\end{linenomath*}
where the sigmoid non-linearity is once again used to obtain values between 0 and 1 for $\vi(t)$, while the candidate update is generated using a $tanh$ non-linearity to ensure that the values in the cell state can be updated both by increasing and decreasing them. In the next step, the cell state is updated based on the vectors produced by the forget
and input gates using element-wise multiplication, denoted by $\odot$. This update is given by
\begin{linenomath*}\begin{align*}
  \vc(t) = \vc(t-1) \odot \vf(t) + \tilde{\vc}(t)\odot \vi(t).
\end{align*}\end{linenomath*}
Then, the next short-term hidden state needs to be generated, which
is based on the now updated cell state. A sigmoid non-linearity is once again used to regulate to which degree the
values of the cell state will be included in the hidden state, and a $tanh$ non-linearity is used to compress the cell
state in the interval (-1, 1). The new hidden state is thus calculated as
\begin{linenomath*}\begin{align*}
  \tilde{\h}(t) &= \sigma(\W_h[\h(t-1), \x(t)] + \vb_h)\\
  \h(t)&= \tilde{\h}(t) \odot tanh(\vc(t)).
\end{align*}\end{linenomath*}
Finally, the output vector is calculated from the hidden state. This part is not always required, as many time
series problems require only one output for a whole sequence. In our case however, we need an output for each time point, which is then
calculated using a simple linear layer as
\begin{linenomath*}\begin{align*}
  \y(t) = \vW_{o}\h(t) + \vb_{o}.
\end{align*}\end{linenomath*}

The interplay of the different gates contributes to the stability of the gradients, making the LSTM very well suited for time
series problems with long-term dependencies.
This is why LSTMs have become one of the most widely used machine learning models for sequential data, such as
speech-to-text recognition~\cite{graves2014towards}, machine translation~\cite{sutskever2014sequence}, video
classification~\cite{donahue2015long}, text~\cite{karpathy2015unreasonable} and image
generation~\cite{oord2016pixel}.

\subsection{Model hyperparameters}\label{ssec:hyperparams}
All of the models used in this paper have certain hyperparameters that need to be optimized and the correct choice of these hyperparameters can in some cases strongly
influence the performance of the model. Of the models used in this paper, the performance of the KRR and ESN models was strongly influenced by the choice of hyperparameters,
while the performance of the other models remained stable over a wider range of hyperparameter values. For the ESN model, which had the largest number of hyperparameters, we used
random search to select sets of hyperparameters for validation, while for the other models the hyperparameters were optimized using a grid search. In the following we report the hyperparameters
that resulted in the lowest validation errors for each of our models.

LRR only has one hyperparameter, the regularization parameter $\lambda$, which was selected as $0.01$ for the synthetic dataset and $0.5$ for the real-world dataset (i.e., the noisier real-world dataset required a stronger protection against outliers compared to the relatively clean synthetic dataset).
KRR has two hyperparameters, the regularization parameter $\lambda$ and the kernel width $\sigma$. The resulting optimal hyperparameters from our cross-validation procedure were $\lambda=0.00046$ and $\sigma=121.99$ for the synthetic dataset and $\lambda=0.00183$ and $\sigma= 1.75$ for the real-world dataset.
Since the other models have multiple hyperparameters that need to be optimized, the optimal values for each dataset are presented in Table~\ref{tab:ffn_hyperparameters} for the FFNN, Table~\ref{tab:esn_hyperparameters} for the ESN model, and Table~\ref{tab:lstm_hyperparameters} for the LSTM model hyperparameters.
\begin{table}[htbp!]
  \centering
  \caption{The optimized hyperparameters for the FFNN model for both datasets.}
  \label{tab:ffn_hyperparameters}
  \begin{tabular}{lcc}
    Parameter & Synthetic dataset &  Real-world dataset\\
    \hline
    Batch size &  128 & 16\\
    Learning rate & 0.000005 & 0.00001 \\
    Momentum & 0.95 & 0.95\\
    Dropout rate & 0 & 0.3\\
    Number of layers & 2 & 2\\
    Layers size & 512 & 512 \\
    \hline
  \end{tabular}
\end{table}
\begin{table}[htbp!]
  \centering
  \caption{The optimized hyperparameters for the ESN model for both datasets.}
  \label{tab:esn_hyperparameters}
  \begin{tabular}{lcc}
    Parameter & Synthetic dataset & Real-world dataset\\
    \hline
    Hidden state size &  343 & 58\\
    Connectivity of $\vW_{rec}$ & 32 & 90 \\
    Distribution non-zero values & Standard normal & Standard normal\\
    Spectral radius of $\vW_{rec}$ & 1.18 & 0.77\\
    Spectral radius of $\vW_{in}$ & 0.004 & 0.0016 \\
    Bias scale for $\vW_{in}$ & 0.68 & 0.08 \\
    Leakage rate & 0.05 & 0.012\\
    \hline
  \end{tabular}
\end{table}
\begin{table}[htbp!]
  \centering
    \caption{The optimized hyperparameters for the LSTM model for both datasets.}
    \label{tab:lstm_hyperparameters}
  \begin{tabular}{lcc}
    Parameter & Synthetic dataset & Real-world dataset\\
    \hline
    Batch size & 64 & 16\\
    Learning rate & 0.001 & 0.005 \\
    Momentum & 0.95 & 0.95\\
    Hidden layer size & 512 & 256 \\
    \hline
  \end{tabular}
\end{table}

\end{document}